\documentclass[lettersize,journal]{IEEEtran}
\usepackage{amsmath,amsfonts}
\usepackage[ruled,linesnumbered]{algorithm2e}

\usepackage{array}
\usepackage[caption=false,font=normalsize,labelfont=sf,textfont=sf]{subfig}
\usepackage{textcomp}
\usepackage{stfloats}
\usepackage{url}
\usepackage{verbatim}
\usepackage{color}
\usepackage{booktabs}
\usepackage{makecell}
\usepackage{cite}
\usepackage{enumerate}
\usepackage{amssymb}
\usepackage{multirow}
\usepackage{comment}
\usepackage{adjustbox}
\usepackage {diagbox}
\usepackage{makecell}

\usepackage{enumitem}

\newcommand{\argmin}{\mathop{\mathrm{arg~min}}\limits}

\begin{document}

% \title{Integrating Feasibility Feedback into Imitation Learning from Observation}
% \title{\textcolor{red}{FABCO: Feasibility-aware Imitation Learning from Observation through Dynamics-based Action and Feasibility Estimation with Multimodal Feedback}
\title{
Feasibility-aware Imitation Learning from Observation \\
with Multimodal Feedback
}

\author{
    Kei Takahashi, 
    Hikaru Sasaki, ~\IEEEmembership{Member,~IEEE}, 
    and Takamitsu Matsubara, ~\IEEEmembership{Member,~IEEE}
            
    %\thanks{Manuscript received April 19, 2021; revised August 16, 2021.}
    \thanks{This work was supported by JSPS KAKENHI Grant Numbers JP24KJ1702 and JP24K03018.}
    \thanks{Kei Takahashi, Hikaru Sasaki, and Takamitsu Matsubara are with the Division of Information Science, Graduate School of Science and Technology, Nara Institute of Science and Technology (NAIST), Ikoma-shi, Nara 630-0192, Japan, (e-mail: takahashi.kei.tl5@is.naist.jp, sasaki.hikaru@is.naist.jp, and takam-m@is.naist.jp).}
}



\maketitle

\begin{abstract}
% Imitation learning frameworks that learn robot control policies from demonstrators' motions via hand-mounted demonstration interfaces have attracted increasing attention.
% However, due to differences in physical characteristics between demonstrators and robots, this approach faces two limitations: i) the demonstration data do not include robot control commands, and ii) the demonstrated motions may be infeasible for robots.
% These limitations make policy learning difficult.
% To address them, we propose Feasibility-Aware Behavior Cloning from Observation (FABCO).
% FABCO integrates behavior cloning from observation, which complements control commands using robot dynamics models, with feasibility estimation.
% In feasibility estimation, the demonstrated motions are evaluated using a robot dynamics model based on reproducibility under the robot's dynamics.
% The estimated feasibility is used for multimodal feedback and feasibility-aware policy learning to improve the demonstrator's motions and learn robust policies.
% Multimodal feedback provides feasibility through the demonstrator's visual and haptic senses to promote feasible demonstrated motions.
% Feasibility-aware policy learning reduces the influence of demonstrated motions that are infeasible for robots, enabling the learning of policies that robots can execute stably.
% We conducted experiments with 15 participants on two tasks and confirmed that FABCO improves imitation learning performance by more than 3.2 times compared to the case without feasibility feedback.
Imitation learning frameworks for learning robot control policies from demonstrators' motions via hand-mounted demonstration interfaces have attracted increasing attention.
However, due to differences in physical characteristics between demonstrators and robots, this approach faces two limitations: i) demonstration data do not include robot actions, and ii) demonstrated motions may be infeasible for robots.
These limitations increase the difficulty of policy learning, so we address them by proposing Feasibility-Aware Behavior Cloning from Observation (FABCO).
FABCO integrates behavior cloning from observation, which complements robot actions using models of robot dynamics, with feasibility estimation.
In feasibility estimation, demonstrated motions are evaluated with a robot-dynamics model, learned from the robot's execution data, to assess reproducibility under the robot's dynamics.
This estimated feasibility is used for multimodal feedback and feasibility-aware policy learning to improve demonstrator motions and learn robust policies.
Multimodal feedback provides feasibility through the demonstrator's visual and haptic senses to promote feasible demonstrated motions.
Feasibility-aware policy learning reduces the influence of demonstrated motions that are infeasible for robots, facilitating the learning of policies that robots can execute stably.
Experiments with 15 participants on two tasks confirm that FABCO improves imitation learning performance by more than 3.2 times compared with learning without feasibility feedback.
\end{abstract}

\def\abstractname{Note to Practitioners}
\begin{abstract}
This study aims to enable non-expert demonstrators without specialized robotics knowledge to intuitively perform robot-feasible motions using a hand-mounted demonstration interface.
In imitation learning, natural demonstrations may include motions that are infeasible for robots, such as excessively quick movements.
If such infeasible motions are used to learn policies,
robots may fail to imitate them, leading to task failure or instability.
To address this issue, we propose Feasibility-Aware Behavior Cloning from Observation (FABCO).
% FABCO estimates the feasibility of each demonstrated motion using robot dynamics models during or after demonstration and provides \textcolor{red}{the} visual \textcolor{red}{feedback system} and \textcolor{red}{the} haptic \textcolor{red}{feedback system} to encourage intuitive corrections.
FABCO estimates the feasibility of each demonstrated motion using robot dynamics models during or after demonstration and provides both visual and haptic feedback to encourage intuitive corrections.
The visual feedback system visualizes the demonstrated motion and color-codes it based on feasibility, allowing the demonstrator to review the motion after performing it.
The haptic feedback system vibrates the handle of the hand-mounted demonstration interface when velocity or pose constraints are exceeded as the motion is being performed, prompting on-the-spot corrections.
These feedback mechanisms enable non-expert demonstrators to perform motions that satisfy the robot's constraints, ensuring safe and practical data collection.
Our approach is useful in settings where tasks change frequently, such as chemical experiments and food processing, by supporting rapid policy updates and improving on-site adaptability.
% In future work, we will develop more intuitive feedback and mechanisms that support automatic correction toward a more general demonstration support system.
\end{abstract}

\begin{IEEEkeywords}
Imitation learning from observation, hand-mounted demonstration interface, feasibility, action chunking and temporal ensembling
% Article submission, IEEE, IEEEtran, journal, \LaTeX, paper, template, typesetting.
\end{IEEEkeywords}

\begin{figure}[t]
    \centering
    \begin{minipage}[b]{0.95\linewidth}
        \centering
        \subfloat[Conventional imitation learning framework \label{fig:intro:overview:prev}]{
        \includegraphics[width=1\hsize]{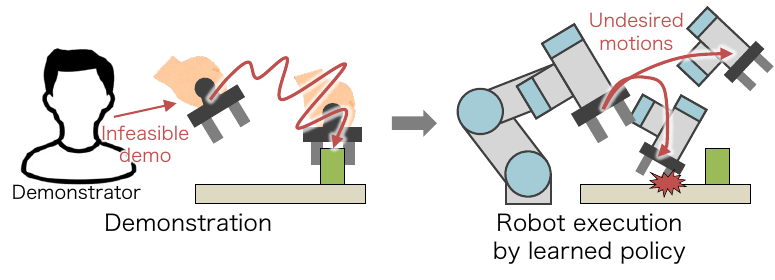}}
    \end{minipage}
    \begin{minipage}[b]{0.95\linewidth}
        \centering
        \subfloat[Proposed imitation learning framework \label{fig:intro:overview:prop}]{
        \includegraphics[width=1\hsize]{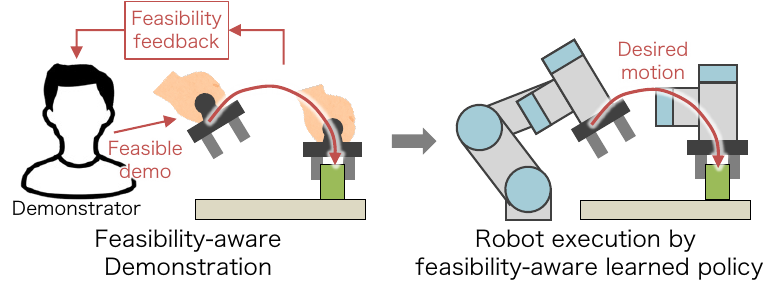}}
    \end{minipage}
    % \caption{Overview of the feasibility-aware demonstration collection method: hand-mounted-interfaceを持って教示した動作に対して，ロボットで実行可能かどうかを判定し，教示者にフィードバックを行う．このサイクルを繰り返すことで教示動作を改善，教示動作の実行可能性の向上を行い，ロボットができる教示動作の収集を行う．}
    \caption{
    Comparison between conventional and proposed feasibility-aware imitation learning frameworks.
    (a) In demonstrations performed with a hand-mounted interface, feasibility constraints are not reflected in demonstrations, resulting in learned policies that may produce unpredictable or undesirable motions.
    (b) Proposed framework incorporates feasibility in both demonstration and learning, promoting feasible demonstrations and improving performance of learned policy toward desired motions.
    }
    \label{fig:intro:overview}
\end{figure}

%%%%%%%%%%%%%%%%%%%%%%%%%%%%%%%%%%%%%%%%%%%%%%%%%%%%%%%%%%%%%%%%%%%%%%%%%%%%%%%%
%%%%%%%%%%%%%%%%%%%%%%%%%%%%%%%%%%%%%%%%%%%%%%%%%%%%%%%%%%%%%%%%%%%%%%%%%%%%%%%%
%%%%%%%%%%%%%%%%%%%%%%%%%%%%%%%%%%%%%%%%%%%%%%%%%%%%%%%%%%%%%%%%%%%%%%%%%%%%%%%%
\section{Introduction}
% Imitation learning acquires control policies for robot automation from demonstrations\cite{osa2018}, and many frameworks have been proposed \cite{chi2024diffusionpolicy,janner2022diffuser,bharadhwaj2024roboagent,schaal1996learning,argall2009survey, torabi2018, baker2022video, torabi2019adversarial}.
% % Conventionally, demonstration data are collected by teleoperating robots, which requires robot expertise.
% Conventionally, demonstration data are collected by teleoperating robots, \textcolor{red}{which often requires expert demonstrators who understand the robot's geometric and motion constraints.}
% Recently, robot-free demonstration approaches that collect demonstration data without robot operation have been actively studied, enabling non-expert demonstrators without specialized robotics knowledge to provide demonstrations.
% In particular, among robot-free demonstration approaches, methods in which a demonstrator wears a hand-mounted demonstration interface to provide demonstrations have attracted attention \cite{WangSWZFL24, chen2024arcap, chi2024universal, hamaya2020, young2021, legato, pmlr-v270-ha25a, zeng2025activeumiroboticmanipulationactive}.

\IEEEPARstart{I}{mitation} learning is an approach to acquiring robot control policies using demonstration data \cite{osa2018}, and many frameworks using it have been proposed \cite{chi2024diffusionpolicy,janner2022diffuser,bharadhwaj2024roboagent,schaal1996learning,argall2009survey, torabi2018, baker2022video, torabi2019adversarial}.
Conventional imitation learning approaches assume that demonstration data are collected through robot teleoperation with an expert robot demonstrator who understands the robot's geometric and motion constraints.
In contrast, recent approaches have explored the strategy of collecting demonstration data from non-expert demonstrators as they perform demonstrations with their own hands.
Among these, demonstration-based approaches using hand-mounted demonstration interfaces have attracted much attention \cite{WangSWZFL24, chen2024arcap, chi2024universal, hamaya2020, young2021, legato, pmlr-v270-ha25a, zeng2025activeumiroboticmanipulationactive}.

However, these demonstration approaches have two limitations because they do not implement a robot in the demonstration:
i) Demonstration data obtained through a hand-mounted demonstration interface do not include data about robot control commands.
ii) Demonstrated motions may not be feasible for real robots due to the different dynamics between human demonstrators and typical robots.
These limitations make it difficult to learn policies for a given task, and the learned policies produce unstable or infeasible motions (Fig. \ref{fig:intro:overview:prev}).

These limitations arise because the robot's dynamics are not used in demonstrations with a hand-mounted demonstration interface.
For teleoperation-based demonstration \cite{wong2022error, fu2024mobile, pmlr-v270-zhao25b, cheng2024opentelevision, cuan2024leveraginghapticfeedbackimprove, Zhao-RSS-23}, the demonstration data are collected as the demonstrator interacts with the robot's dynamics during robot operation.
A teleoperation-based demonstration not only allows the collection of robot actions but also constrains demonstrated motions to feasible robot motions.
However, a hand-mounted interface-based demonstration does not benefit from the advantages of a teleoperation-based demonstration.

We aim to achieve an imitation learning framework that enables even novice demonstrators, without expert knowledge of robots, to learn appropriate policies (Fig. \ref{fig:intro:overview:prop}).
Therefore, we introduce a robot's dynamics into imitation learning using a hand-mounted demonstration interface with the intention of predicting the robot's actions and thus improving the feasibility of the demonstrator's motion.

In this study, we propose Feasibility-Aware Behavior Cloning from Observation (FABCO) as an imitation learning framework incorporating a hand-mounted demonstration interface.
FABCO is based on behavior cloning from observation (BCO) \cite{torabi2018}, an imitation learning method that uses a robot dynamics model to supplement action in demonstration data, and it uses the feasibility estimated by this model for both demonstration and policy learning.
The estimated feasibility is fed back multimodally to the demonstrator's visual and tactile senses to guide feasible motion demonstration, and it is used for policy learning based on feasible demonstration data.
Here, feasibility is estimated using both the robot's forward dynamics model (FDM) and its inverse dynamics model (IDM).

Preliminary results of using FABCO were presented previously 
\cite{takahashi2025feasibilityawareimitationlearningobservations}.
Building upon and significantly extending this preliminary work, this study extends FABCO by enhancing both the feasibility feedback and the policy model, as well as by conducting more comprehensive experimental evaluations.
The feasibility feedback is designed to present feasibility both during and after demonstrations to the demonstrator, guiding them in revising and improving their motions.
Specifically, in addition to the conventional visual feedback that presents feasibility information as images after a demonstration, we introduce haptic feedback to convey feasibility as vibrations during the demonstration.
These two feedback methods thus provide true multimodal feedback.
The performance of policies learned by FABCO depends heavily on the accuracy of the inverse dynamics model (IDM), and policy performance can degrade significantly due to the IDM's modeling errors.
To achieve policy learning that is robust to modeling error, we incorporate action chunking and temporal ensemble (ACTE) \cite{Zhao-RSS-23} into the policy model.
Experiments were conducted with 15 participants and two tasks having distinct characteristics to clarify the generality and reproducibility of FABCO.
To evaluate the workload experienced by demonstrators during demonstration with feedback, we administered a questionnaire using the NASA Task Load Index (NASA-TLX) \cite{HART1988139}.

The contributions of this work can be summarized as follows:
\begin{itemize}
    \item We propose FABCO, a framework that estimates demonstrated motion feasibility and integrates it into multimodal feedback and policy learning.
    \item We clarify the advantages and disadvantages of each of two feedback methods based on their characteristics and the task characteristics.
    \item We evaluate the effectiveness and generality of FABCO through experiments with two robot tasks and 15 subjects.
\end{itemize}

%%%%%%%%%%%%%%%%%%%%%%%%%%%%%%%%%%%%%%%%%%%%%%%%%%%%%%%%%%%%%%%%%%%%%%%%%%%%%%%%
%%%%%%%%%%%%%%%%%%%%%%%%%%%%%%%%%%%%%%%%%%%%%%%%%%%%%%%%%%%%%%%%%%%%%%%%%%%%%%%%
%%%%%%%%%%%%%%%%%%%%%%%%%%%%%%%%%%%%%%%%%%%%%%%%%%%%%%%%%%%%%%%%%%%%%%%%%%%%%%%%
\section{Related Work}
\subsection{Demonstration methods in imitation learning}
In imitation learning, previous studies have reported demonstration methods for collecting data conveniently to learn imitation policies.
Representative demonstration methods can be categorized into two types: teleoperation-facilitating demonstrations and robot-free demonstrations.
Among those of the first type, various teleoperation methods have been attempted \cite{wong2022error, fu2024mobile, pmlr-v270-zhao25b, cheng2024opentelevision, cuan2024leveraginghapticfeedbackimprove, Zhao-RSS-23}.
Wong et al.\ proposed a system for operating a mobile robot while checking images from an onboard camera via a smartphone \cite{wong2022error}.
As a state-of-the-art leader--follower demonstration approach, Mobile Aloha has been proposed and widely used, facilitating demonstrations not only for a robot hand but also for base locomotion \cite{fu2024mobile}.
A key advantage of teleoperation-facilitating demonstrations is the ability to collect robot-feasible demonstration data, since the robot physically executes motions during the demonstration.
For robot-free demonstrations, demonstration data are collected without the use of a robot.
Among the robot-free demonstrations, in some approaches humans demonstrate motions with their own hands \cite{liu2018imitation, smith2019avid, qin2021from, zhang2022learning, bahl2022human, karnan2022voila, sikchi2024dual}, while in others humans hold a hand-mounted demonstration interface to demonstrate and learn a policy from the collected data \cite{WangSWZFL24, chen2024arcap, chi2024universal, hamaya2020, young2021, legato, pmlr-v270-ha25a, zeng2025activeumiroboticmanipulationactive}.
Qin et al.\ built a system that collects data by capturing human hand motions with a camera and converting them into robot hand motions in a simulator \cite{qin2021from}.
Chi et al.\ designed a hand-mounted demonstration interface that shares the end-effector structure with robots, permitting data collection and policy learning from intuitively demonstrated motion\cite{chi2024universal}.
% A key advantage of robot-free demonstrations is that robots are not required during data collection, which makes demonstrations intuitive to provide.
A key advantage of the robot-free type is that demonstrators can provide demonstrations by moving their own hands or a hand-mounted demonstration interface without operating a robot, which makes the demonstration process intuitive.

% \textcolor{red}{Conventional methods have not achieved both easy demonstration data collection and demonstrations that are guaranteed to be executable by robots.
% Teleoperation-facilitating demonstrations can yield robot-executable data, but still require operating the robot during data collection.
% Robot-free demonstrations enable easy data collection, but may include demonstrations that are infeasible for robots.}
Conventional methods have achieved neither intuitive collection of demonstrated motions nor any guarantee that robots can execute them.
In this study, we construct a method that succeeds in both of these complementary goals.
Specifically, a hand-mounted demonstration interface enables demonstrators to provide demonstrated motions without robot operation, while dynamics-based feasibility estimation and feedback support the demonstration of motions that real robots can execute.

\subsection{Applying feasibility to imitation learning}
Using feasibility for imitation learning makes a policy more likely to generate stable motions, leading to higher task success rates and improved safety.
Therefore, recent studies have increasingly focused on leveraging robot feasibility in imitation learning.

Feasibility-aware imitation learning methods can be broadly grouped into two directions.
One direction provides feasibility feedback to the demonstrator during data collection to promote the collection of highly feasible demonstration data\cite{chen2024arcap, sukkar2023guided}.
Chen et al.\ defined feasibility as satisfying robot constraints, such as workspace and velocity limits, and provided visual feedback by presenting these constraints via AR as the demonstrated motions are performed \cite{chen2024arcap}.
Their results show improved task success rates for learned policies, showing the effectiveness of feasibility feedback during demonstration.
The other direction uses feasibility during policy learning \cite{cao2021corl, cao2021ral, Betz_2021}.
Cao et al.\ considered a cross-robot transfer setting, where demonstration data collected on a high-DOF robot are applied to a lower-DOF robot.
In this setting, feasibility is defined as cross-robot imitability, and their method learns a policy by discarding low-feasibility data during training \cite{cao2021corl, cao2021ral}.

Providing feasibility feedback during motion demonstration can improve the quality of the demonstrated motions, but policy performance can still degrade if low-feasibility demonstration data remain in the policy learning data.
In contrast, incorporating feasibility during policy learning improves robustness by excluding low-feasibility data, but it may reduce the effective data available for policy learning and thus risk decreased policy performance.
To address these limitations, this study constructs a learning framework that leverages feasibility in both demonstration and policy learning, improving demonstrated motion quality and the quality of the learned policy.
Specifically, FABCO collects highly feasible demonstration data through feasibility feedback during motion demonstration and performs data selection based on the estimated feasibility for policy learning.

\section{Preliminaries}

%%%%%%%%%%%%%%%%%%%%%%%%%%%%%%%%%%%%%%%%%%%%%%%%%%%%%%%%%%%%%%%%%%%%%%%%%%%%%%%%
\subsection{Behavior Cloning from Observation}
% BCO is an imitation learning method that learns a policy from state-only demonstration data \cite{torabi2018}.
% Because actions are not available in these demonstrated motions, BCO includes a step that learns an IDM and uses it to infer actions from state transitions for policy learning.

BCO learns a control policy from only the states observed in a demonstration \cite{torabi2018}.
Specifically, to compensate for the lack of action in the demonstration data, BCO learns an inverse dynamics model (IDM) of the robot and predicts the robot's actions from the observed states to learn the policy.

%%%%%%%%%%%%%%%%%%%%%%%%%%%%%%%%%%%%%%%%%%%%%%%%%%%%%%%%%%%%%%%%%%%%%%%%%%%%%%%%
\subsubsection{Problem setting}
% BCO assumes a Markov decision process. The initial state $\mathbf s_1$ follows $p(\mathbf s_1)$, and given a state $\mathbf s_t$ and an action $\mathbf a_t$ at time $t$, the next state $\mathbf s_{t+1}$ follows $p(\mathbf s_{t+1}\mid \mathbf s_t, \mathbf a_t)$.
% Let the demonstrator policy and the agent policy be $\pi^e(\mathbf a_t\mid \mathbf s_t)$ and $\pi^r_\phi(\mathbf a_t\mid \mathbf s_t)$, respectively.
% The demonstrator acts according to $\pi^e(\mathbf a_t\mid \mathbf s_t)$, and we assume that only the demonstrated state sequence $\mathbf S^e = \{\mathbf s_t^e\}_{t=1}^T$ is available for learning the agent policy.

BCO assumes a Markov decision process.
The initial state $\mathbf s_1$ follows initial state probability $p(\mathbf s_1)$, and given a state $\mathbf s_t$ and an action $\mathbf a_t$ at time $t$, the next state $\mathbf s_{t+1}$ follows state transition probability $p(\mathbf s_{t+1}\mid \mathbf s_t, \mathbf a_t)$.
Let the policies of the demonstrator and the robot be $\pi^e(\mathbf a_t\mid \mathbf s_t)$ and $\pi^r_\phi(\mathbf a_t\mid \mathbf s_t)$, respectively.
The demonstrated motions are generated according to the policy $\pi^e(\mathbf a_t\mid \mathbf s_t)$, and we assume that only the demonstrated state sequence $\mathbf S^e = \{\mathbf s_t^e\}_{t=1}^T$ is available for policy learning.

%%%%%%%%%%%%%%%%%%%%%%%%%%%%%%%%%%%%%%%%%%%%%%%%%%%%%%%%%%%%%%%%%%%%%%%%%%%%%%%%
\subsubsection{IDM learning and action estimation}
% BCO uses an IDM to estimate actions from the demonstrated state sequence $\mathbf S^e$.
% To obtain an IDM, BCO collects an agent dataset by executing an arbitrary tracking policy $\pi^\mathrm{track}$ in the environment, yielding a state sequence $\mathbf S^r = \{\mathbf s_t^r\}_{t=1}^T$ and an action sequence $\mathbf A^r = \{\mathbf a_t^r\}_{t=1}^T$.
% The IDM $f^\mathrm{IDM}_\theta$ predicts an action $\mathbf a_t$ from a pair of consecutive states $\mathbf s_t$ and  $\mathbf s_{t+1}$.
% Assuming that the collected trajectories $(\mathbf S^r, \mathbf A^r)$ are sampled from the distribution induced by executing $\pi^\mathrm{track}$, denoted by $p(\mathbf S^r, \mathbf A^r\mid \pi^\mathrm{track})$,
% the IDM parameters $\theta$ are optimized as :
% \begin{align}
%     \theta^* &= \argmin_\theta J_\theta, \\
%     J_\theta &= \mathbb E_{p(\mathbf S^r, \mathbf A^r\mid \pi^\mathrm{track})}\left[
%     \sum_{t=1}^{T-1} \left\| \mathbf a_t^r - f^\mathrm{IDM}_\theta(\mathbf s_t^r, \mathbf s_{t+1}^r) \right\|_1
%     \right].
% \end{align}
% With the optimized parameters $\theta^*$, actions for the demonstration are estimated as
% $\tilde{\mathbf a}^e_t = f^\mathrm{IDM}_{\theta^*}(\mathbf s^e_t,\mathbf s^e_{t+1})$.

To obtain an IDM, BCO collects a state sequence $\mathbf S^r = \{\mathbf s_t^r\}_{t=1}^T$ and an action sequence $\mathbf A^r = \{\mathbf a_t^r\}_{t=1}^T$ by executing a robot according to an arbitrary tracking policy.
The IDM $f^\mathrm{IDM}_\theta$ is a model that predicts an action $\mathbf a_t$ from a pair of consecutive states $\mathbf s_t$ and  $\mathbf s_{t+1}$.
Assuming that the collected state and action sequence $\mathbf S^r$, $\mathbf A^r$ are sampled from the distribution $p(\mathbf S^r, \mathbf A^r\mid \pi^\mathrm{track})$,
the IDM parameters $\theta$ are optimized as
\begin{align}
    \theta^* &= \argmin_\theta J_\theta, \\
    J_\theta &= \mathbb E_{p(\mathbf S^r, \mathbf A^r\mid \pi^\mathrm{track})}\left[
    \sum_{t=1}^{T-1} \left\| \mathbf a_t^r - f^\mathrm{IDM}_\theta(\mathbf s_t^r, \mathbf s_{t+1}^r) \right\|_1
    \right].
\end{align}
With the optimized parameters $\theta^*$, actions for the demonstration are estimated as
$\tilde{\mathbf a}^e_t = f^\mathrm{IDM}_{\theta^*}(\mathbf s^e_t,\mathbf s^e_{t+1})$.

%%%%%%%%%%%%%%%%%%%%%%%%%%%%%%%%%%%%%%%%%%%%%%%%%%%%%%%%%%%%%%%%%%%%%%%%%%%%%%%%

\subsubsection{Policy learning}
The agent policy $\pi_\phi^r$ is learned from the demonstration state sequence $\mathbf S^e$ using the IDM.
Assuming $\mathbf S^e=\{\mathbf s_t^e\}_{t=1}^T$ is generated by the demonstrator policy $\pi^e$ and follows $p(\mathbf S^e\mid \pi^e)$, the policy parameters $\phi$ are optimized as
\begin{align}
    \phi^* &= \argmin_\phi J_\phi, \\
    J_\phi &= \mathbb E_{p(\mathbf S^e\mid \pi^e)}\left[
    \sum_{t=1}^{T-1}
    \left\|
    f^\mathrm{IDM}_{\theta^*}(\mathbf s^e_t,\mathbf s^e_{t+1}) - \pi_\phi^r (\mathbf s^e_t)
    \right\|_1
    \right].
\end{align}

%%%%%%%%%%%%%%%%%%%%%%%%%%%%%%%%%%%%%%%%%%%%%%%%%%%%%%%%%%%%%%%%%%%%%%%%%%%%%%%%
%%%%%%%%%%%%%%%%%%%%%%%%%%%%%%%%%%%%%%%%%%%%%%%%%%%%%%%%%%%%%%%%%%%%%%%%%%%%%%%%
\subsection{Action Chunking and Temporal Ensembling}
% ACTE \cite{Zhao-RSS-23} combines action chunking and temporal ensembling to mitigate compounding errors in imitation learning.
ACTE\cite{Zhao-RSS-23} is a method that combines action chunking, which treats actions as a short-horizon sequence, and temporal ensembling, which integrates chunked action predictions over time.
The use of ACTE is expected to mitigate compounding errors, a common issue in imitation learning.
In action chunking, the policy predicts an action chunk of length $k$ at time $t$,
$\mathbf a_t^{\mathrm{ch}} = [\mathbf a_{t \mid t}, \cdots, \mathbf a_{t+k\mid t}]$,
as
$\mathbf a_t^{\mathrm{ch}} = \pi_\phi^{\mathrm{ch}}(\mathbf s_t)$,
where $\mathbf a_{t'\mid t}$ denotes the action at time $t'$ predicted at time $t$.

Temporal ensembling $\mathrm{TE}(\cdot,\cdot)$ determines the executed action $\mathbf a_t$ by taking a weighted average of the past $k$ action predictions with weights $\mathbf w^\mathrm{ch} = [w_1^\mathrm{ch}, \cdots, w_{k}^\mathrm{ch}]$:
\begin{align}
    \mathbf a_t = \mathrm{TE}(\mathbf w^\mathrm{ch}, \mathbf a_{t:t-k}^\mathrm{ch})
    = \frac{\sum_{i=1}^{k} w_i^\mathrm{ch} \mathbf a_{t\mid t-i}}{\sum_{i=1}^{k} w_i^\mathrm{ch}}.
    \label{eq:temporal_ensemble}
\end{align}
Here, the weights $\mathbf w^\mathrm{ch}$ are manually designed in advance.

%%%%%%%%%%%%%%%%%%%%%%%%%%%%%%%%%%%%%%%%%%%%%%%%%%%%%%%%%%%%%%%%%%%%%%%%%%%%%%%%
%%%%%%%%%%%%%%%%%%%%%%%%%%%%%%%%%%%%%%%%%%%%%%%%%%%%%%%%%%%%%%%%%%%%%%%%%%%%%%%%

\section{Feasibility-Aware Behavior Cloning from Observation}
\begin{figure}[!t]
    \centering
    \begin{minipage}[b]{0.8\linewidth}
        \centering
        \subfloat[Demonstration with feasibility feedback \label{fig:proposed:overview:demo}]{
        \includegraphics[width=\hsize]{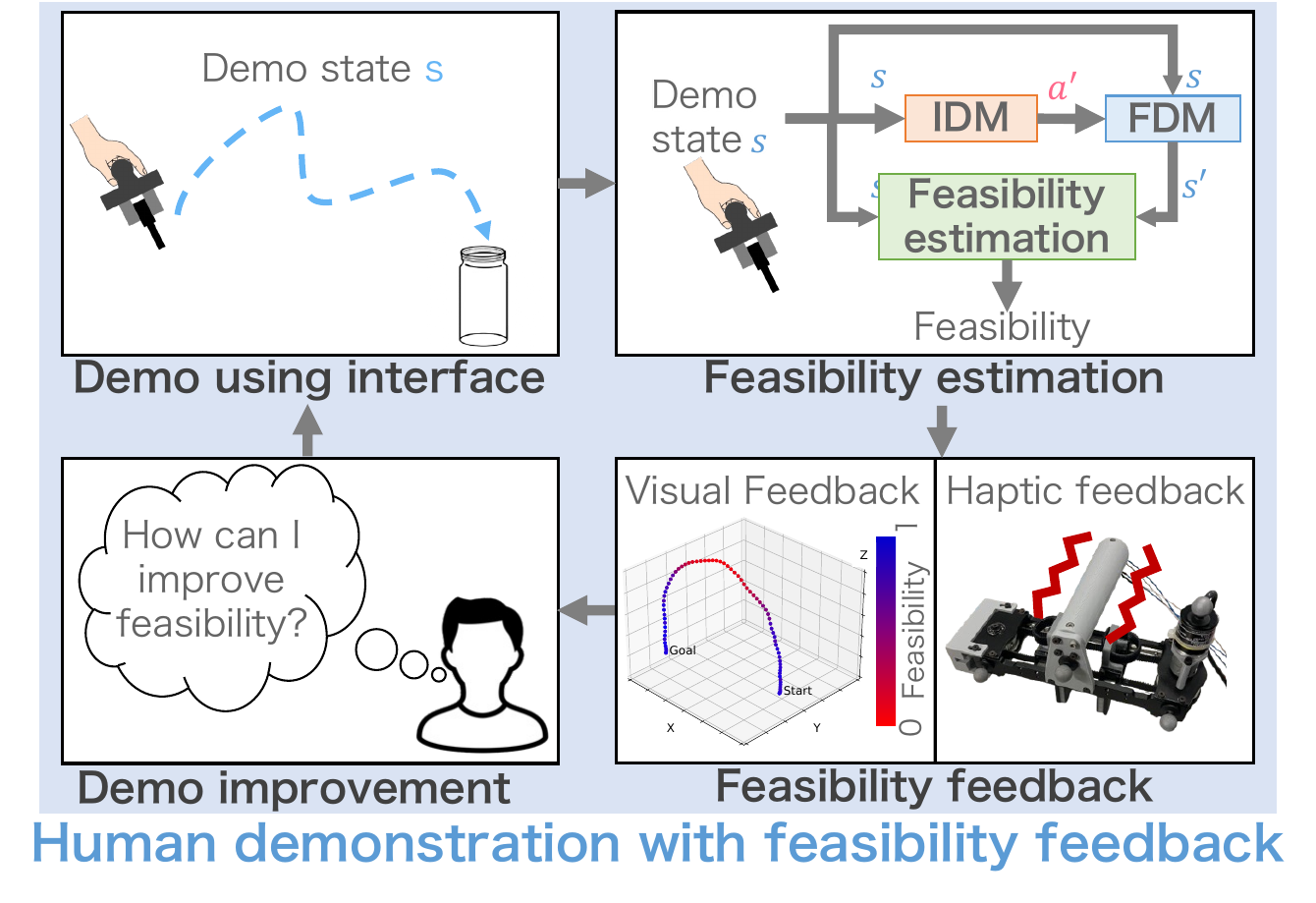}}
    \end{minipage}
    \begin{minipage}[b]{0.8\linewidth}
        \centering
        \subfloat[Feasibility-aware policy learning \label{fig:proposed:overview:learn}]{
        \includegraphics[width=1\hsize]{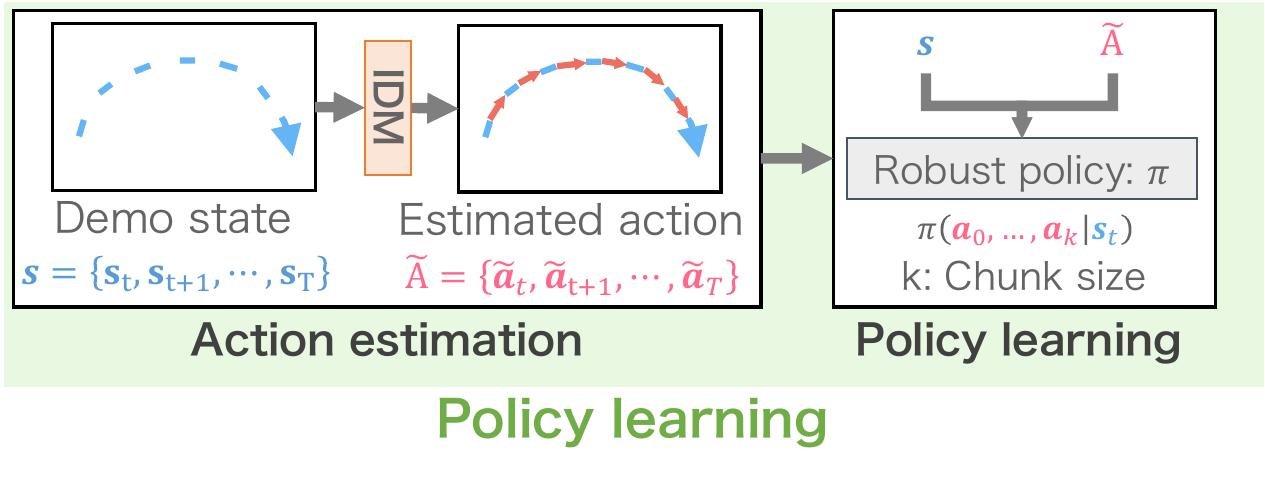}}
    \end{minipage}
    \caption{
    Overview of FABCO framework.
    This framework incorporates feasibility into both demonstration and policy learning in imitation learning using a hand-mounted demonstration interface.
    (a) The demonstrator receives feasibility feedback on demonstrated motions, which is computed using FDM and IDM, and refines the demonstration based on this feedback.
    (b) Policy learning uses the feasibility of the demonstration data, permitting the learned policy to generate feasible robot motions.
    }
    \label{fig:proposed:overview}
\end{figure}

In this section, we introduce FABCO, an imitation learning framework that provides feasibility feedback and learns a policy based on feasibility (Fig.~\ref{fig:proposed:overview}).
FABCO is a learning algorithm based on BCO that estimates feasibility using a robot's IDM and FDM learned in advance and that leverages feasibility for both demonstration and policy learning.

%%%%%%%%%%%%%%%%%%%%%%%%%%%%%%%%%%%%%%%%%%%%%%%%%%%%%%%%%%%%%%%%%%%%%%%%%%%%%%%%
%%%%%%%%%%%%%%%%%%%%%%%%%%%%%%%%%%%%%%%%%%%%%%%%%%%%%%%%%%%%%%%%%%%%%%%%%%%%%%%%
\subsection{Problem Statement}
A state $\mathbf s_t$ in the state sequence $\mathbf S=\{\mathbf s_t\}_{t=1}^T$ consists of the robot's end-effector pose $\mathbf p_t$ and an environmental observation $\mathbf o_t$.
The end-effector pose is represented by a six-dimensional vector of position and orientation,
$\mathbf p_t = \{p_{t,x}, p_{t,y}, p_{t,z}, p_{t,\mathrm{roll}}, p_{t,\mathrm{pitch}}, p_{t,\mathrm{yaw}}\}$.
To learn an FDM and an IDM, we assume that robot actions can be inferred from transitions in the end-effector pose sequence $\{\mathbf p_t\}_{t=1}^T$.
Based on this assumption, the IDM predicts the action from end-effector pose transitions, while the FDM predicts the next end-effector pose given the current pose and action.
For training the FDM and IDM, random reference trajectories are prepared within the robot workspace, and the robot follows them using a tracking policy $\pi^\mathrm{track}$.
During trajectory tracking, we collect a pose sequence $\mathbf P^r=\{\mathbf p^r_t\}_{t=1}^T$ and an action sequence $\mathbf A^r=\{\mathbf a^r_t\}_{t=1}^T$.
For policy learning, we infer actions from pose transitions in the demonstration data using the learned IDM.

%%%%%%%%%%%%%%%%%%%%%%%%%%%%%%%%%%%%%%%%%%%%%%%%%%%%%%%%%%%%%%%%%%%%%%%%%%%%%%%%
%%%%%%%%%%%%%%%%%%%%%%%%%%%%%%%%%%%%%%%%%%%%%%%%%%%%%%%%%%%%%%%%%%%%%%%%%%%%%%%%
\subsection{Demonstration interface}
\begin{figure}[t]
    \centering
    \subfloat[Overview \label{fig:experiment:common_device:overview}]{\includegraphics[width=0.4\linewidth]{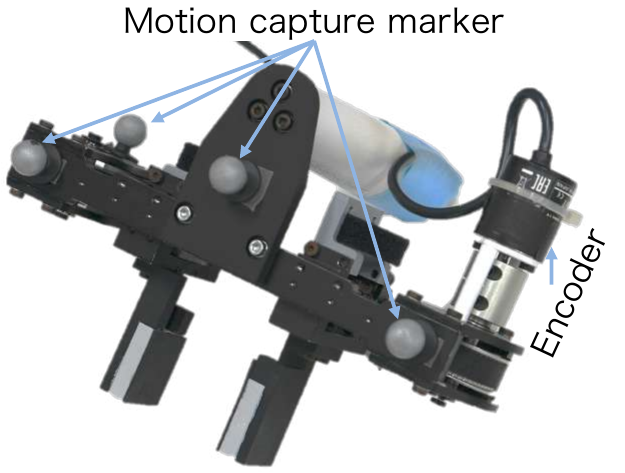}}
    % \hfill
    \subfloat[Internal structure \label{fig:experiment:common_device:motor}]{\includegraphics[width=0.4\linewidth]{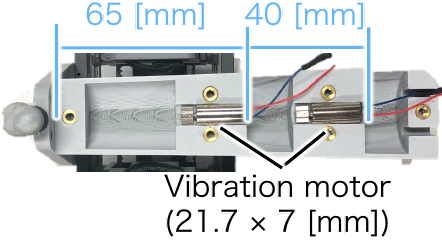}}
    \caption{Hand-mounted demonstration interface. 
    (a) Overview of interface. An encoder is attached to measure finger position. Motion-capture markers are attached to track the interface pose.
    (b) Internal structure of handle. It is equipped with two vibration motors.}
    \label{fig:experiment:common_device}
\end{figure}

To collect demonstration data with an end-effector structure shared between the robot and the human, we developed a hand-mounted demonstration interface that matches a two-fingered parallel robot hand (Fig.~\ref{fig:experiment:common_device}).
This interface is inspired by the work of Hamaya et al.\ \cite{hamaya2020}, and it allows a demonstrator to open and close the fingers by manipulating a handle.
The finger opening width is measured by an encoder attached to the interface, and the interface pose is tracked using a motion capture system.
To provide feasibility feedback, we mounted two cylindrical vibration motors (KPD7C-0716) on the interface handle (Fig.~\ref{fig:experiment:common_device:motor}).
Each motor has a rated speed of 7500 rpm, and the vibration intensity is controlled by pulse-width modulation (PWM) of the two motors.

%%%%%%%%%%%%%%%%%%%%%%%%%%%%%%%%%%%%%%%%%%%%%%%%%%%%%%%%%%%%%%%%%%%%%%%%%%%%%%%%
%%%%%%%%%%%%%%%%%%%%%%%%%%%%%%%%%%%%%%%%%%%%%%%%%%%%%%%%%%%%%%%%%%%%%%%%%%%%%%%%
\subsection{Learning of forward and inverse dynamics models}
We learn the FDM and IDM from robot motion data collected while the robot tracks randomly prepared reference trajectories in the workspace.
For this data collection, trajectories are prepared within the robot workspace, and the robot follows them using a tracking policy $\pi^\mathrm{track}$.
During trajectory tracking, we collect a pose sequence $\mathbf P^r$ and an action sequence $\mathbf A^r$.
We assume that these pose and action sequences are distributed according to $p(\mathbf P^r, \mathbf A^r \mid \pi^\mathrm{track})$.

The IDM $f_\theta^\mathrm{IDM}(\mathbf p_t, \mathbf p_{t+1})$ takes consecutive robot poses $\mathbf p_t$ and $\mathbf p_{t+1}$ as input and predicts the action $\mathbf a_t$ that induces this pose transition.
The FDM $f_\psi^\mathrm{FDM}(\mathbf p_t, \mathbf a_t)$ predicts the next pose $\mathbf p_{t+1}$ from the current pose $\mathbf p_t$ and action $\mathbf a_t$.
Let $\theta$ and $\psi$ denote the parameters of the IDM and FDM, respectively.
These are optimized as
\begin{align}
    \theta^* &= \argmin_\theta J_\theta, \\
    J_\theta &= \mathbb E_{p(\mathbf P^r,\mathbf A^r\mid \pi^\mathrm{track})}\left[
    \sum_{t=1}^{T-1} \left\| \mathbf a_t^r - f^\mathrm{IDM}_\theta(\mathbf p_t^r, \mathbf p_{t+1}^r)\right\|_1
    \right], \label{eq:IDM} \\
    \psi^* &= \argmin_\psi J_\psi, \\
    J_\psi &= \mathbb E_{p(\mathbf P^r,\mathbf A^r\mid \pi^\mathrm{track})}\left[
    \sum_{t=1}^{T-1} \left\| \mathbf p_{t+1}^r - f^\mathrm{FDM}_\psi(\mathbf p_t^r, \mathbf a_t^r)\right\|_1
    \right]. \label{eq:FDM}
\end{align}
With the optimized parameters $\theta^*$ and $\psi^*$, the IDM predicts the action $\tilde{\mathbf a}_t = f_{\theta^*}^\mathrm{IDM}(\mathbf p_t, \mathbf p_{t+1})$, and the FDM predicts the next pose $\tilde{\mathbf p}_{t+1} = f_{\psi^*}^\mathrm{FDM}(\mathbf p_t, \tilde{\mathbf a}_t)$.

%FDMを先に言って，IDMを後に言うように変えた
% The FDM $f_\psi^\mathrm{FDM}(\mathbf p_t, \mathbf a_t)$ predicts the next pose $\mathbf p_{t+1}$ from the current pose $\mathbf p_t$ and action $\mathbf a_t$.
% The IDM $f_\theta^\mathrm{IDM}(\mathbf p_t, \mathbf p_{t+1})$ takes consecutive robot poses $\mathbf p_t$ and $\mathbf p_{t+1}$ as input and predicts the action $\mathbf a_t$ that induces this pose transition.
% Let $\psi$ and $\theta$ denote the parameters of the FDM and IDM, respectively.
% They are optimized as:
% \begin{align}
%     \psi^* &= \argmin_\psi J_\psi, \\
%     J_\psi &= \mathbb E_{p(\mathbf P^r,\mathbf A^r\mid \pi^\mathrm{track})}\left[
%     \sum_{t=1}^{T-1} \left\| \mathbf p_{t+1}^r - f^\mathrm{FDM}_\psi(\mathbf p_t^r, \mathbf a_t^r)\right\|_1
%     \right], \label{eq:FDM} \\
%     \theta^* &= \argmin_\theta J_\theta, \\
%     J_\theta &= \mathbb E_{p(\mathbf P^r,\mathbf A^r\mid \pi^\mathrm{track})}\left[
%     \sum_{t=1}^{T-1} \left\| \mathbf a_t^r - f^\mathrm{IDM}_\theta(\mathbf p_t^r, \mathbf p_{t+1}^r)\right\|_1
%     \right]. \label{eq:IDM}
% \end{align}
% With the optimized parameters $\psi^*$ and $\theta^*$, the IDM predicts the action $\tilde{\mathbf a}_t = f_{\theta^*}^\mathrm{IDM}(\mathbf p_t, \mathbf p_{t+1})$, and the FDM predicts the next pose $\tilde{\mathbf p}_{t+1} = f_{\psi^*}^\mathrm{FDM}(\mathbf p_t, \tilde{\mathbf a}_t)$.

%%%%%%%%%%%%%%%%%%%%%%%%%%%%%%%%%%%%%%%%%%%%%%%%%%%%%%%%%%%%%%%%%%%%%%%%%%%%%%%%
%%%%%%%%%%%%%%%%%%%%%%%%%%%%%%%%%%%%%%%%%%%%%%%%%%%%%%%%%%%%%%%%%%%%%%%%%%%%%%%%
\subsection{Feasibility estimation}
Feasibility is estimated from the end-effector pose sequence $\mathbf P^e=\{\mathbf p^e_t\}_{t=1}^T$ in the demonstration data $\mathbf S^e$ collected with the hand-mounted demonstration interface, together with the IDM and FDM.
% The IDM estimates an action $\tilde{\mathbf a}^e_t = f^\mathrm{IDM}_{\theta^*}(\mathbf p_t^e, \mathbf p_{t+1}^e)$, and the FDM predicts the resulting pose transition $\tilde{\mathbf p}^e_{t+1} = f^\mathrm{FDM}_{\psi^*}( \mathbf p_t^e, \tilde{\mathbf a}^e_t)$.
Using the IDM and FDM, feasibility is defined as
\begin{align}
    f_{t,d} &= \exp
        \left(
            -\frac{\mid\tilde p^e_{t+1,d} - p^e_{t+1,d}\mid}{\sigma_f}
        \right),
    \label{eq:feasibility:pose-level} \\
    f_t &= \frac{1}{6}\sum_{d\in \mathcal P} f_{t,d},
    \label{eq:feasibility:overall}
\end{align}
where $f_{t,d}$ denotes the feasibility for the $d$-th pose component, $f_t$ denotes the overall feasibility at time step $t$, $\tilde p^e_{t+1,d}$ and $p^e_{t+1,d}$ denote the $d$-th dimensional component of the predicted end-effector pose and the demonstrated end-effector pose at time $t+1$, respectively, $\sigma_f$ controls the scale of feasibility, and $\mathcal P=\{x,y,z,\mathrm{roll},\mathrm{pitch},\mathrm{yaw}\}$ is the set of pose components.
By using the exponential function, $f_{t,d}$ and $f_t$ are represented as values that smoothly vary between 0 and 1.

\subsection{Multimodal feasibility feedback}
Multimodal feasibility feedback encourages demonstrators to provide motions that robots can readily imitate.
We employ two types of feedback: visual feedback and haptic feedback.
The visual feedback presents the feasibility of an entire demonstrated motion after the motion is performed,
whereas the haptic feedback evaluates feasibility as the motion is performed and delivers immediate feedback by vibrating the handle of the hand-mounted demonstration interface.

The visual feedback provides offline feasibility information by displaying the end-effector position sequence $\mathbf P^e$ on a screen and color-coding it according to feasibility.
Specifically, this visualization uses $f_{t,d}$ computed by Eq.~\ref{eq:feasibility:pose-level} and $f_t$ computed by Eq.~\ref{eq:feasibility:overall} to show pose-level and overall feasibility, respectively.

The haptic feedback system evaluates feasibility for the end-effector pose $\mathbf p^e_t$ at time step $t$ and provides online feedback by driving vibration motors mounted on the hand-mounted demonstration interface (Fig.~\ref{fig:experiment:common_device}).
Using the overall feasibility $f_t$ computed by Eq.~\ref{eq:feasibility:overall}, the PWM signal $s_\mathrm{PWM}$ is computed as
\begin{align}
    s_\mathrm{PWM} =
    \begin{cases}
        0     & (1-f_t < \tau), \\
        1-f_t & (\text{otherwise}),
    \end{cases}
\end{align}
% where $\tau$ denotes the threshold for providing haptic feasibility feedback.
where $\tau$ denotes the threshold for the haptic feedback to generate vibration.
Accordingly, no vibration is generated when feasibility is sufficiently high, and the vibration intensity increases as feasibility decreases.

%%%%%%%%%%%%%%%%%%%%%%%%%%%%%%%%%%%%%%%%%%%%%%%%%%%%%%%%%%%%%%%%%%%%%%%%%%%%%
%%%%%%%%%%%%%%%%%%%%%%%%%%%%%%%%%%%%%%%%%%%%%%%%%%%%%%%%%%%%%%%%%%%%%%%%%%%%%%%%
\subsection{Feasibility-aware policy learning}
For policy learning, we use the demonstrated motion $\mathbf S^e =\{\mathbf s_t^e\}_{t=1}^T$, the feasibility  $\mathbf F^e =\{\mathbf f_t\}_{t=1}^T$ for the demonstrated motion obtained via the dynamics models, and the estimated actions $\tilde{\mathbf A}^e =\{\tilde{\mathbf a}_t\}_{t=1}^T$.
To learn a policy robustly from low-feasibility demonstrated data, we extend the objective function using weights $\mathbf W^e = \{w_t^e\}_{t=1}^T$ for policy learning.

To compute the weights used for policy learning, we use a step-wise evaluation $e^\mathrm{step}_t$ based on feasibility and an episode-wise evaluation $e^\mathrm{episode}$.
The step-wise evaluation is computed as follows:
\begin{align}
    e^\mathrm{step}_t =
    \begin{cases}
        1 & (f_t \ge \kappa), \\
        0 & (\text{otherwise}),
    \end{cases}
\end{align}
where $\kappa$ is the step-wise threshold that determines whether the demonstration data at time $t$ is used for learning.
The episode-wise evaluation uses the step-wise evaluations and prohibits using an episode when low-feasibility data continues for a certain duration.
It is computed using the following function:
\begin{align}
    e^\mathrm{episode} =
    \begin{cases}
        0 & \left(f_\mathrm{MaxRun}\left([e^\mathrm{step}_t]_{t=1}^T\right) \ge \gamma\right), \\
        1 & (\text{otherwise}),
    \end{cases}
\end{align}
where $f_\mathrm{MaxRun}$ returns the maximum length of consecutive zeros in the step-wise evaluations $e_t^{\mathrm{step}}$ and $\gamma$ is the episode-wise threshold that determines whether to use the entire episode for learning.
The weight for the demonstration data at each time step is computed as the product of the step-wise evaluation $e^\mathrm{step}_t$ and the episode-wise evaluation $e^\mathrm{episode}$:
\begin{align}
    w^e_t =  e^\mathrm{episode} \cdot e^\mathrm{step}_t.
\end{align}

For policy learning, we designed a feasibility-weighted objective function to implement feasibility-aware policy learning.
The policy model uses action chunking and outputs an action sequence up to $k$ steps ahead, and it is estimated by minimizing the following feasibility-weighted objective:
\begin{align}
    \phi^*  &= \argmin_\phi  \, J_\phi^\mathrm{ch}, \label{eq:policy2} \\
    J_{\phi}^\mathrm{ch}
    &= \mathbb E_{p(\mathbf S^e\mid \pi^e)}\left[
        \sum_{t=1}^{T - k}
        \left\|
            \mathbf w^e_t \odot
            \{\tilde{\mathbf a}_{t}^\mathrm{ch} - \pi^\mathrm{ch}_\phi(\mathbf s^e_{t})\}
        \right\|
    \right],
    \label{eq:policy1}
\end{align}
% \begin{align}
%     \phi^*  &= \argmin_\phi  \, J_\phi^\mathrm{ch}, \\
%     J_{\phi}^\mathrm{ch}
%     &= \mathbb E_{p(\mathbf S^e\mid \pi^e)}\left[
%         \sum_{t=1}^{T - k}
%         \textcolor{red}{w^e_t} \,
%         \left\|
%             \tilde{\mathbf a}_{t}^\mathrm{ch} - \pi^\mathrm{ch}_\phi(\mathbf s^e_{t})
%         \right\|_1
%     \right],
% \end{align}
where $\odot$ denotes the Hadamard product, $\tilde{\mathbf a}_t^\mathrm{ch} = [\tilde{\mathbf a}_{t+i}]_{i=0}^{k-1}$ denotes a $k$-step chunked action vector by predicted actions $\tilde{\mathbf a}_t$, and $\mathbf w_t^e = [w_i^e]_{i=t}^{t+k}$ denotes the weight vector corresponding to the chunked action vector.
When the learned policy $\pi_{\phi^*}^\mathrm{ch}$ is applied, the action $\mathbf {a_t} $ is determined using temporal ensembling as given in  Eq. \ref{eq:temporal_ensemble}.

% \begin{align}
%     \phi^*  &= \argmin_\phi \, J_\phi^{\mathrm{ch}}, \label{eq:policy2}\\
%     \color{red}
%     J_\phi^{\mathrm{ch}}
%     &= \color{red}
%        \mathbb E_{p(\mathbf S^e\mid \pi^e)}\left[
%         \sum_{t=1}^{T-k}
%         \sum_{i=0}^{k-1}
%         w_{t+i}^e \,
%         \left\|
%             \tilde{\mathbf a}_{t+i} - \pi_{\phi}^{\mathrm{ch}}(\mathbf s_t^e)
%         \right\|_1
%     \right],
%     \color{black}
%     \label{eq:policy1}
% \end{align}
% where $\tilde{\mathbf a}_{t+i}$ is the estimated action at time step $t+i$, and $w_{t+i}^e$ is the corresponding  weight.
% The chunking policy $\pi_{\phi}^{\mathrm{ch}}(\mathbf s_t^e)$ outputs a $k$-step action sequence given the state $\mathbf s_t^e$.

% \color{red}
% To determine the executed action $\mathbf a_t$, we use temporal ensembling.
% At each time step, the policy produces a $k$-step action sequence, and multiple recent sequences provide overlapping predictions for the same time step.
% We compute the executed action $\mathbf a_t$ as a weighted combination of these overlapping predictions using temporal-ensemble weights $\mathbf w^{\mathrm{ch}}$.
% \color{black}

%%%%%%%%%%%%%%%%%%%%%%%%%%%%%%%%%%%%%%%%%%%%%%%%%%%%%%%%%%%%%%%%%%%%%%%%%%%%%%%%
%%%%%%%%%%%%%%%%%%%%%%%%%%%%%%%%%%%%%%%%%%%%%%%%%%%%%%%%%%%%%%%%%%%%%%%%%%%%%%%%
\subsection{FABCO's learning algorithm}
FABCO is an imitation learning framework that estimates the feasibility of demonstrated motions using a robot's FDM and IDM, encourages demonstrators to provide robot-feasible motions by feeding back this feasibility to the demonstrator, and finally uses the feasibility for policy learning.
The learning algorithm of FABCO is shown in Algorithm~\ref{alg:Learning_process}.

\begin{algorithm}[t]
    % \footnotesize
    \small
    \SetAlgoLined
    \DontPrintSemicolon
    % \SetKwInOut{Input}{Input}
    % \SetKwInOut{Output}{Output}
    % \Input{State $\mathbf S^r = \{\mathbf P^r ,\mathbf O\}$}
    % \Output{Action $\mathbf A^r$}
    %0204修正
    % Initialize FDM $f_{\theta}^\mathrm{FDM}$, IDM $f_{\psi}^\mathrm{IDM}$ and , policy $\pi_{\phi}$, {chunk size $k$}\;
    % {Let $\mathbf s_t$, $\mathbf a_t$ represent action and state at timestep $t$}\;
    Initialize IDM $f_{\theta}^\mathrm{IDM}$, FDM $f_{\psi}^\mathrm{FDM}$, policy $\pi_{\phi}$, and chunk size $k$\;
    Let $\mathbf s_t$ and $\mathbf a_t$ denote the state and action at time step $t$, respectively\;
    \textbf{\textit {\# Learning process of FDM and IDM}} \;
    Generate robot's trajectories $\mathcal D^r=\{\mathbf S^r_n, \mathbf A^r_n\}_{n=1}^N$ using trajectory tracking policy $\pi^\mathrm{track}$ \;    
    % Improve $f_{\theta}^\mathrm{IDM}$ by modelLearning\{($\mathbf p^r_t, \mathbf p^r_{t+1}), (\mathbf a^r_t)\}_{t=1}^T$ \;
    % Improve $f_{\phi}^\mathrm{FDM}$ by modelLearning\{($\mathbf p^r_t, \mathbf a^r_t), (\mathbf p^r_{t+1})\}_{t=1}^T$ \;
    Learn IDM $f_{\theta}^\mathrm{IDM}$ using data $\mathcal D$ by Eq. (\ref{eq:IDM})  \;
    Learn FDM $f_{\psi}^\mathrm{FDM}$ using data $\mathcal D$ by Eq. (\ref{eq:FDM})  \;
    \textbf{\textit {\# Feasibility-aware demonstration process}} \;
    \For{$m = 1$ \textbf{to} $M$}{
        Collect demonstrated motion $\mathbf S^e_m$\;
        Predict action $\tilde{\mathbf A}^e_m$ for robot pose $\mathbf P^e_m$ using IDM\;
        % Use FDM with ($\mathbf P^e_m$, $\tilde{\mathbf A}^e_m$) to approximate $\tilde{\mathbf P}^e_m$\;
        Predict the transition of robot pose $\tilde{\mathbf P}^e_m$ for predicted action $\tilde{\mathbf A}^e_m$ and robot pose $\mathbf P^e_m$ using FDM\;
        % Use FDM with ($\mathbf P^e_m$, $\tilde{\mathbf A}^e_m$) to approximate $\tilde{\mathbf P}^e_m$\;
        Calculate feasibility $\mathbf F^e_m$ using $\mathbf P^e_m$ and $\tilde{\mathbf P}^e_m$ by Eqs. (\ref{eq:feasibility:pose-level}) and (\ref{eq:feasibility:overall})\;
        Feasibility feedback $\mathbf F^e_m$ to the demonstrator\;
    }
    \textbf{\textit{\# Feasibility-aware policy learning }}\;
    Learn policy $\pi_{\phi}$ using data $\mathcal D_\pi = \{\mathbf S^e_m$, $\tilde{\mathbf A}^e_m, \mathbf F^e_m\}_{m=1}^M$ that consists of demonstration data and predicted actions and feasibilities by Eq. (\ref{eq:policy1}) \;
    \caption{Learning process of FABCO}
    \label{alg:Learning_process}
\end{algorithm}

%%%%%%%%%%%%%%%%%%%%%%%%%%%%%%%%%%%%%%%%%%%%%%%%%%%%%%%%%%%%%%%%%%%%%%%%%%%%%%%%
%%%%%%%%%%%%%%%%%%%%%%%%%%%%%%%%%%%%%%%%%%%%%%%%%%%%%%%%%%%%%%%%%%%%%%%%%%%%%%%%
%%%%%%%%%%%%%%%%%%%%%%%%%%%%%%%%%%%%%%%%%%%%%%%%%%%%%%%%%%%%%%%%%%%%%%%%%%%%%%%%
\section{Experiments}
To evaluate FABCO, we conducted a human-subject experiment with an actual robot and 15 subjects.
As a baseline, we used BCO, which does not use feasibility in either demonstrated motion feedback or policy learning.

We tackled two tasks, a peg-insertion task requiring precise position control and a circle-tracing task requiring velocity control.
Using the collected demonstration data and the learned policies, we investigated the following:
\begin{enumerate}[label=RQ\arabic*:, leftmargin=2.5em]
    \item Does feasibility feedback improve the feasibility of demonstrated motions?\label{feasibility}
    \item How does each feedback method affect the demonstrator's workload?\label{workload}
    \item Does the demonstration data collected with feasibility feedback improve policy performance?\label{policy}
    \item To what extent does feasibility-aware policy learning improve policy performance?\label{weighting}
\end{enumerate}

%%%%%%%%%%%%%%%%%%%%%%%%%%%%%%%%%%%%%%%%%%%%%%%%%%%%%%%%%%%%%%%%%%%%%%%%%%%%%%%%
\subsection{Experimental settings}

%%%%%%%%%%%%%%%%%%%%%%%%%%%%%%%%%%%%%%%%%%%%%%%%%%%%%%%%%%%%%%%%%%%%%%%%%%%%%%%%
\subsubsection{Comparison methods}
In this experiment, we evaluated the effectiveness of feasibility feedback and feasibility-aware policy learning by comparing FABCO with the baseline method BCO.
In BCO, demonstration data are collected under the no-feedback (No FB) demonstration, and policies are learned without feasibility.
FABCO collects demonstration data by three demonstration methods: visuo-haptic feedback demonstration, which presents feasibility using both visual and haptic feedback systems; visual feedback demonstration, which presents feasibility using only the visual feedback system; and haptic feedback demonstration, which presents feasibility using only the haptic feedback system.
Using the demonstration data collected under each demonstration method, we compared the feasibility of the demonstrated motion and the performance of the learned policies.
The threshold for triggering the haptic feedback was set to $\tau=0.7$.

%%%%%%%%%%%%%%%%%%%%%%%%%%%%%%%%%%%%%%%%%%%%%%%%%%%%%%%%%%%%%%%%%%%%%%%%%%%%%%%%
\subsubsection{Experimental Procedure}
We recruited 15 subjects and randomly assigned them to one of three demonstration methods (visual, haptic, or visuo-haptic), with five subjects per demonstration method.
Each subject performed both tasks under two demonstration settings: No FB and the assigned demonstration method.
Before each task, subjects received instructions on the task setup, the hand-mounted demonstration interface, and the feedback method, followed by one minute of practice.
For all subjects, the No FB demonstration was performed first, followed by demonstrations under the assigned demonstration method.
After each task was carried out by each demonstration method, subjects completed questionnaires on workload and method preference.

%%%%%%%%%%%%%%%%%%%%%%%%%%%%%%%%%%%%%%%%%%%%%%%%%%%%%%%%%%%%%%%%%%%%%%%%%%%%%%%%
\subsubsection{Verification method}
We conducted human-subject experiments and policy learning experiments on two tasks to investigate RQ1--RQ4.
All the experiments were conducted under the approval of the institutional ethics committee.

For RQ1, we computed the feasibility of each subject's demonstrated motions and evaluated how much each feedback method improved the demonstrated motion's feasibility.
For the peg-insertion task, each subject performed 30 demonstrated motions under each demonstration method.
For the circle-tracing task, each subject performed eight demonstrated motions under each demonstration method, and each demonstrated motion consisted of tracing a circle for five laps.

For RQ2, we evaluated subjects' workload and preferences using NASA-TLX and a preference questionnaire.
In the preference questionnaire, subjects who performed the visuo-haptic feedback demonstration rated ``Which is more beneficial, the visual feedback system or haptic feedback system?'' on a 20-point scale, and subjects who performed the visual or visuo-haptic feedback demonstration rated ``Which is more beneficial, pose-level visualization or overall visualization?'' also on a 20-point scale.

For RQ3, we learned a policy for each subject using the demonstration data collected for RQ1 under each feedback method, and we compared policy performance using robot task success rates.
For the peg-insertion task, we conduct 10 trials per policy, resulting in 150 total trials for No FB and 50 total trials for each of the other feedback methods.
For the circle-tracing task, we conducted five trials per policy, where each trial consists of tracing five circles, resulting in 75 total trials for No FB and 25 total trials for each of the other feedback methods.

For RQ4, we investigated the impact of feasibility-aware policy learning by comparing performance with and without feasibility-aware policy learning.
Using demonstration data collected in the visuo-haptic feedback demonstration, we considered two learning conditions: with full data and first-half data.
In the full data condition, each policy was learned from all demonstrated motions, which were 30 motions for the peg-insertion task and eight motions for the circle-tracing task.
In the first-half data condition, each policy was learned using only the demonstrated motions collected in the first half of the data collection.
For each learning condition, we executed the task with policies learned with and without feasibility-aware policy learning and compared their success rates and Hausdorff distances to the demonstration data.

%%%%%%%%%%%%%%%%%%%%%%%%%%%%%%%%%%%%%%%%%%%%%%%%%%%%%%%%%%%%%%%%%%%%%%%%%%%%%%%%
\subsubsection{Hardware}
% The experimental environment is shown in Fig.~\ref{fig:experiment:environment_all}.
% The hardware devices of the manipulator, the motion-capture system, the robot hand, and the controller have parameters of control frequency and acceleration that are the same as those in our preliminary study~\cite{takahashi2025feasibilityawareimitationlearningobservations}.
% Compared to the preliminary study, however, we used an improved hand-mounted demonstration interface (Fig.~\ref{fig:experiment:common_device}) to implement the haptic feedback.

\begin{figure}[t]
    \centering
    \begin{minipage}[b]{0.5\linewidth}
        \centering
        \subfloat[Experimental environment \label{fig:experiment:environment}]{
        \includegraphics[width=\hsize]{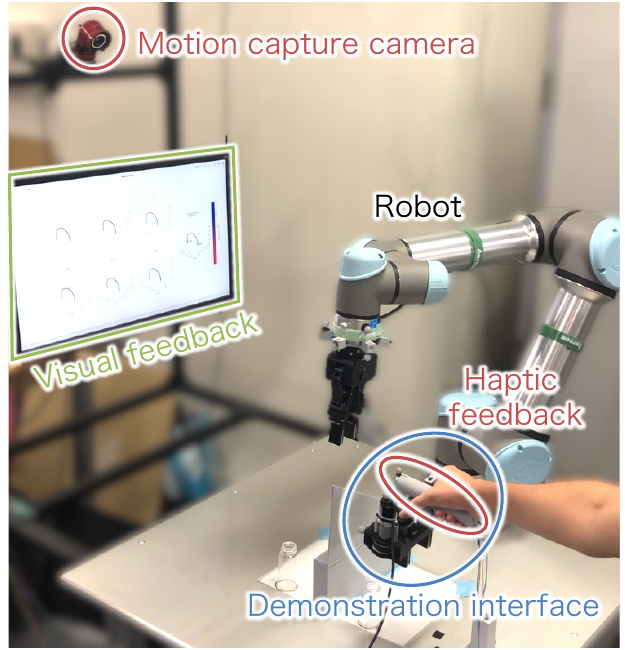}
        }
    \end{minipage}
    \begin{minipage}[b]{0.4\linewidth}
        \centering
        \subfloat[Visual feasibility feedback\label{fig:experiment:feasibility_visualization}]{
            \includegraphics[width=\hsize]{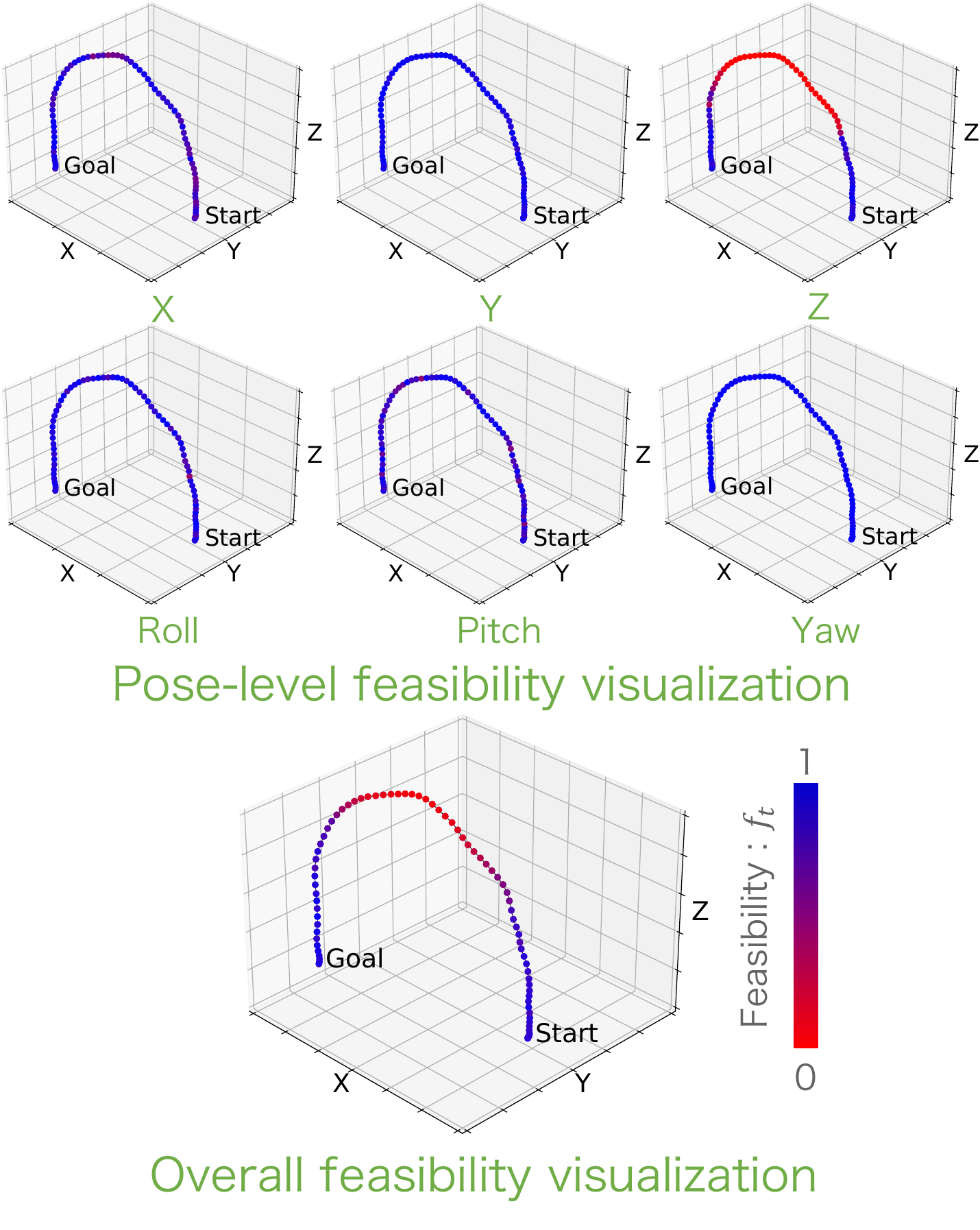}
        }
    \end{minipage}
    \caption{
    Experimental setup for human-subject experiment.
    % (a) \textcolor{red}{During demonstration, the monitor displays feasibility visualizations provided by the visual feedback system.}
    (a) Experimental environment:
    In this environment, the workspace on the desk is shared with a demonstrator and the robot. The motion is captured by a motion-capture camera. The demonstrator receives the feasibility through the monitor and the interface.
    (b) Visual feasibility feedback system: This system provides two types of visualizations: pose-level feasibility visualization and overall feasibility visualization.}
    \label{fig:experiment:environment_all}
\end{figure}

The experimental environment is shown in Fig.~\ref{fig:experiment:environment_all}.
The hardware devices of the manipulator, the motion-capture system, the robot hand, and the controller
use the same control-frequency and acceleration settings as those in the preliminary study
~\cite{takahashi2025feasibilityawareimitationlearningobservations}.
Compared to the preliminary study, however, we used an improved hand-mounted demonstration interface (Fig.~\ref{fig:experiment:common_device})
to implement haptic feedback.

%%%%%%%%%%%%%%%%%%%%%%%%%%%%%%%%%%%%%%%%%%%%%%%%%%%%%%%%%%%%%%%%%%%%%%%%%%%%%%%%
\subsubsection{Peg-insertion task}
\begin{figure}[t]
    \centering
    \begin{minipage}[b]{0.46\linewidth}
        \centering
        \subfloat[Peg-insertion task\label{fig:experiment:pipette_insertion_environment}]{
            \includegraphics[width=\hsize]{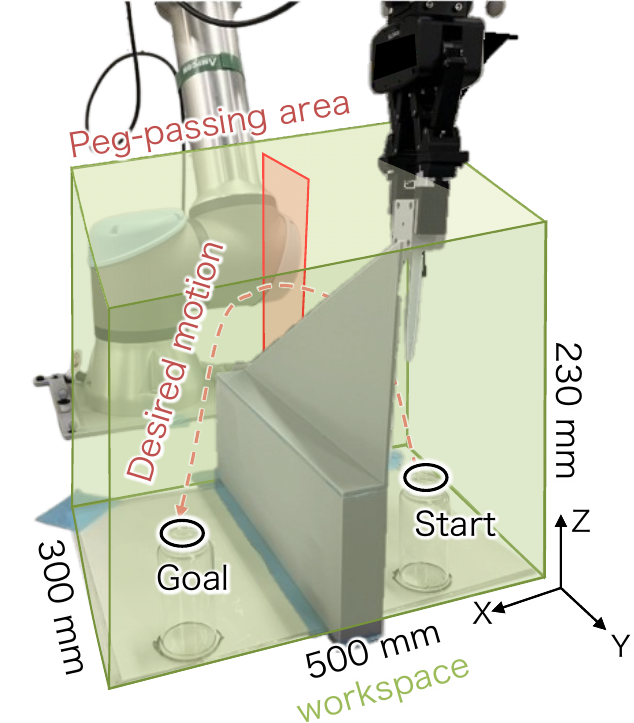}
        }
    \end{minipage}
    \begin{minipage}[b]{0.46\linewidth}
        \centering
        \subfloat[Circle-tracing task\label{fig:experiment:circle_tracing_environment}]{
            \includegraphics[width=\hsize]{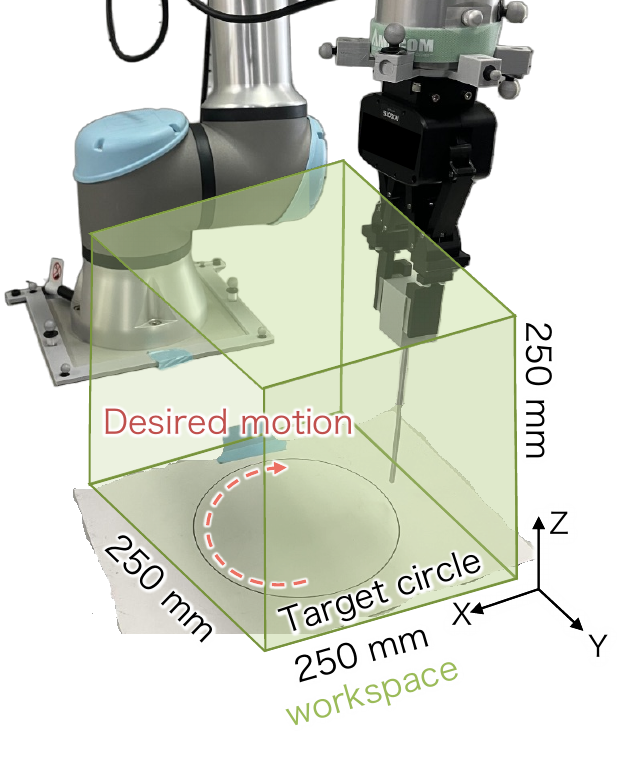}
        }
    \end{minipage}
    \caption{
    Task environments.
    % (a) The peg insertion task moves a grasped peg from start to goal while avoiding obstacles, \textcolor{red}{and must be completed without leaving the green $500\,\mathrm{mm}\times300\,\mathrm{mm}\times230\,\mathrm{mm}$ region.}
    % (b) The circle tracing task traces a circular path on the table \textcolor{red}{and must be completed without leaving the green $250\,\mathrm{mm}\times250\,\mathrm{mm}\times250\,\mathrm{mm}$ region.}
    (a) The peg-insertion task requires moving a grasped peg from start to goal while avoiding obstacles. 
    (b) The circle-tracing task requires tracing a circular path on a table.
    In each task, the green region denotes the workspace in which the robot must complete motions.
    }
\end{figure}

The environment for the peg insertion task is shown in Fig.~\ref{fig:experiment:pipette_insertion_environment}.
The goal of the task is to transfer a peg from the vial on the right to the vial on the left while avoiding the obstacle.
The peg is 130~mm long with a 3~mm $\times$ 3~mm square cross-section, and it has a 30~mm $\times$ 30~mm $\times$ 40~mm rectangular grasping block attached to the top.
The state is the six-dimensional end-effector pose $\mathbf{s}_t=\mathbf{p}_t$, and the action is the six-dimensional target end-effector velocity $\mathbf{a}_t$.
The relative positioning of the peg and the end-effector remains fixed during demonstration and execution.
We define task success as inserting the peg without deviating from the workspace and without contacting the obstacle.
Successful demonstrated motions must pass through the red region in Fig.~\ref{fig:experiment:pipette_insertion_environment}.
This task is designed to evaluate whether FABCO can suppress demonstrated motions that deviate from the workspace.

%%%%%%%%%%%%%%%%%%%%%%%%%%%%%%%%%%%%%%%%%%%%%%%%%%%%%%%%%%%%%%%%%%%%%%%%%%%%%%%%
\subsubsection{Circle-tracing task}
The environment for the circle tracing task is shown in Fig.~\ref{fig:experiment:circle_tracing_environment}.
We define success in the task as completing five laps while tracing the target circle on the table with the tip of the grasped peg, keeping the Hausdorff distance~\cite{jungeblut2022complexityhausdorffdistance} to the demonstrated motions  within 30~mm.
The definitions of the state and action are the same as those in the peg-insertion task.
The relative positioning of the peg and the end-effector also remains fixed during demonstration and execution.
This task evaluates how effectively each feedback method suppresses demonstrated motions that exceed the robot's end-effector speed limit, which is a primary factor reducing feasibility in this task.

%%%%%%%%%%%%%%%%%%%%%%%%%%%%%%%%%%%%%%%%%%%%%%%%%%%%%%%%%%%%%%%%%%%%%%%%%%%%%%%%
\subsubsection{Learning}
The FDM and IDM were modeled by neural networks with five fully connected layers and then learned using data collected as the robot tracked random reference trajectories within each task workspace.
Random reference trajectories were prepared by sampling waypoints in the workspace and connecting them with spline trajectories.
The FDM estimates the next end-effector pose $\mathbf p_{t+1}$ from the current end-effector pose $\mathbf p_t$ and action $\mathbf a_t$.
In this experiment, the FDM is implemented as six independent models that separately predict the six components of $\mathbf p_{t+1}$.
The IDM takes the end-effector poses at three time steps, from $t-1$ to $t+1$, as input and estimates the action $\mathbf a_t$ at time $t$.

The policy is also modeled by a neural network with five fully connected layers.
For ACTE, we set the action chunk length to $k=8$ and use weights $\mathbf w^{\mathrm{ch}} = [8,7,6,5,4,3,2,1]$.
For feasibility-aware policy learning, we set the step-wise threshold $\kappa=0.7$ and the episode-wise threshold $\gamma=10$.

%%%%%%%%%%%%%%%%%%%%%%%%%%%%%%%%%%%%%%%%%%%%%%%%%%%%%%%%%%%%%%%%%%%%%%%%%%%%%%%%
%%%%%%%%%%%%%%%%%%%%%%%%%%%%%%%%%%%%%%%%%%%%%%%%%%%%%%%%%%%%%%%%%%%%%%%%%%%%%%%%
\subsection{Experimental results}
%%%%%%%%%%%%%%%%%%%%%%%%%%%%%%%%%%%%%%%%%%%%%%%%%%%%%%%%%%%%%%%%%%%%%%%%%%%%%%%%
\subsubsection{Improving demonstrated motions via feasibility feedback}
\label{experiment:demo_change}

\begin{figure}[t]
    \centering
    \includegraphics[width=0.7\hsize]{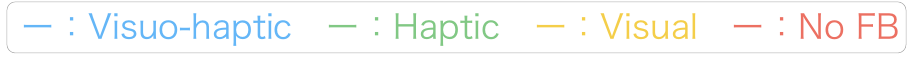}
    \begin{minipage}{0.8\linewidth}
        \centering
        \subfloat[Peg-insertion task \label{fig:experiment:pipette_insertion_feasibility}]{
            \includegraphics[width=\hsize]{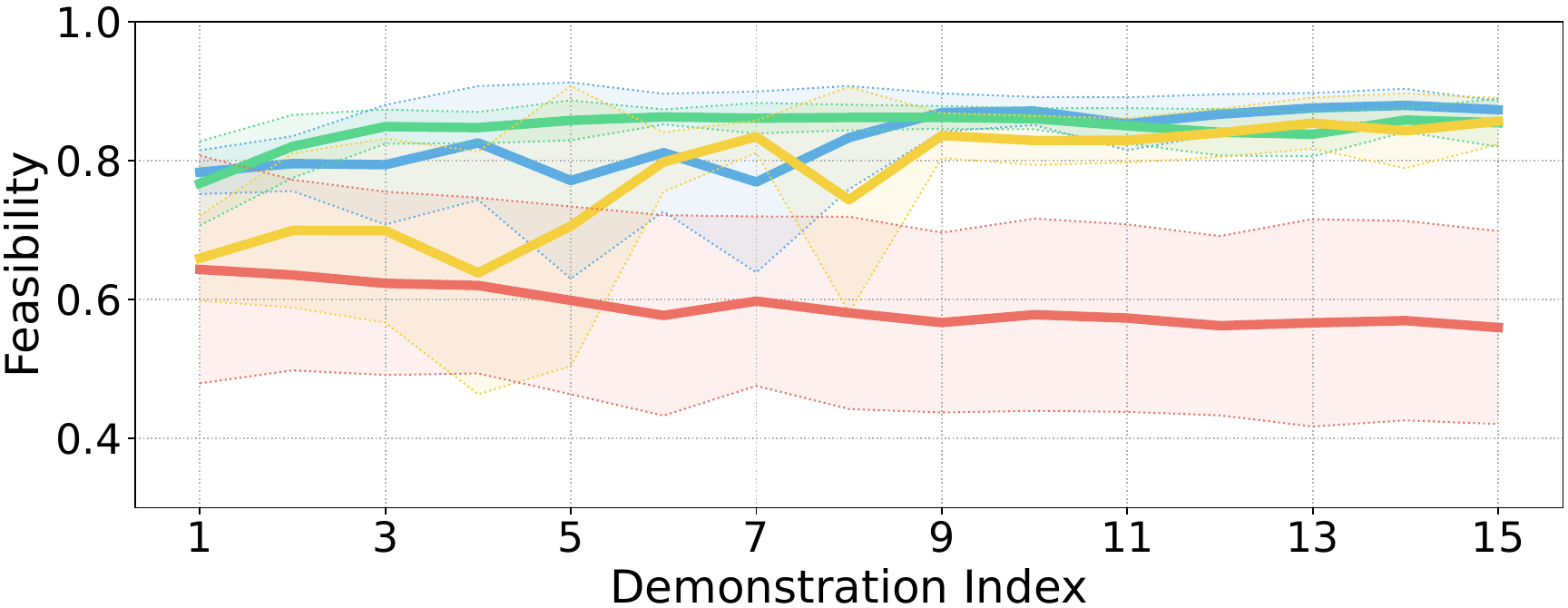}
        } 
    \end{minipage}
    \begin{minipage}{0.8\linewidth}
        \centering
        \subfloat[Circle-tracing task \label{fig:experiment:circle_tracing_feasibility}]{
            \includegraphics[width=\hsize]{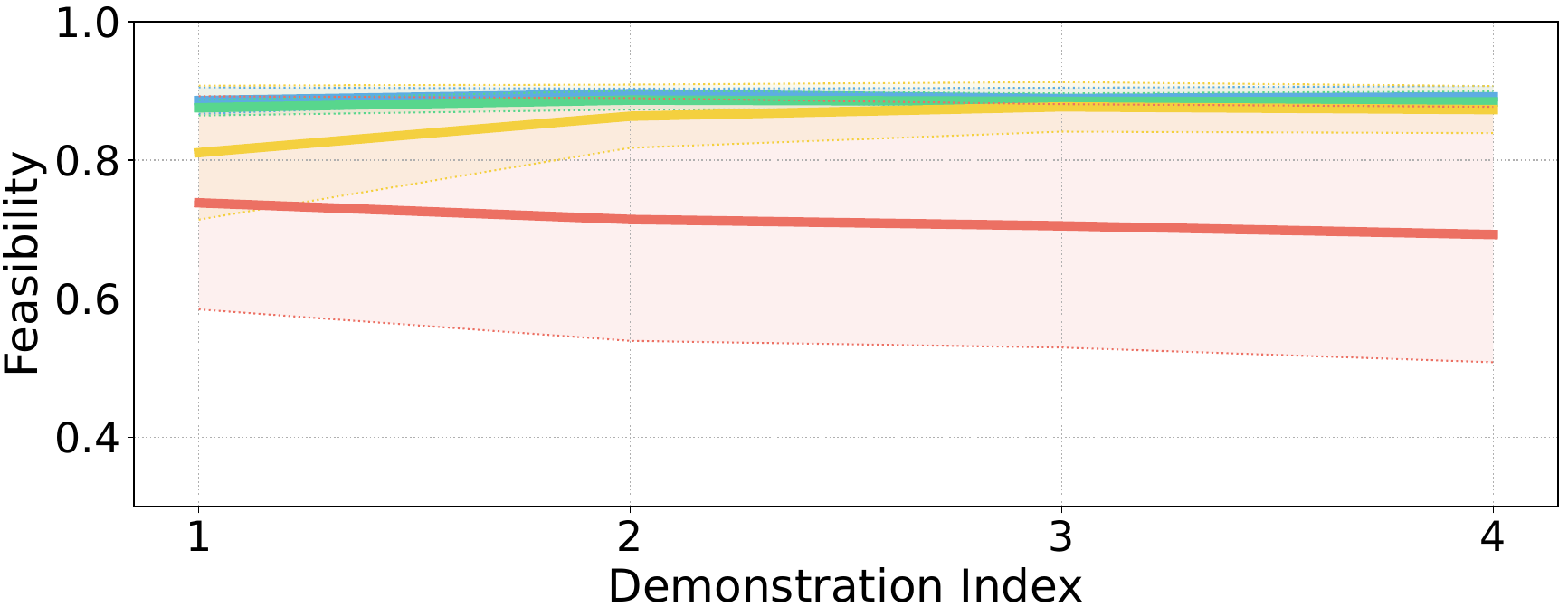}
        }
    \end{minipage}
    % \caption{
    % Transitions of estimated feasibility during motion demonstration.
    % (a) shows the transition in the peg insertion task, and (b) shows the transition in the circle tracing task.
    % The horizontal axis represents the number of demonstrated motions, and the vertical axis represents the mean feasibility of a demonstrated motions.
    % Each plot shows the mean and standard deviation of feasibility transitions across subjects.
    % }
    \caption{
    Transitions of estimated feasibility during demonstration:
    (a) transition in the peg-insertion task and (b) transition in the circle-tracing task.
   Horizontal axis represents number of demonstrated motions, and vertical axis represents mean feasibility of demonstrated motions.
    Each plot shows mean and standard deviation of feasibility transitions.
    For No FB, statistics are computed across a single experiment by 15 subjects.
    For the other methods, statistics are computed across a single experiment by five subjects each.
    % \textcolor{red}{The mean feasibility for each demonstration method is computed from demonstrated motions provided by 15 subjects, all of whom performed the No FB demonstration and were additionally assigned to one of three feedback groups (Visual, Haptic, or Visuo-haptic), with five subjects per feedback group.}
    }
    \label{fig:experiment:feasibility}
\end{figure}
\begin{figure}[t]
    \centering
    \includegraphics[width=0.8\hsize]{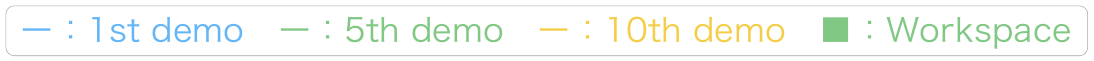}
    % First row
    \begin{minipage}[b]{0.46\linewidth}
        \centering
        \subfloat[\label{fig:experiment:visuo-haptic_trajectory}Visuo-haptic demo \label{fig:a}]{
        \includegraphics[width=1\hsize]{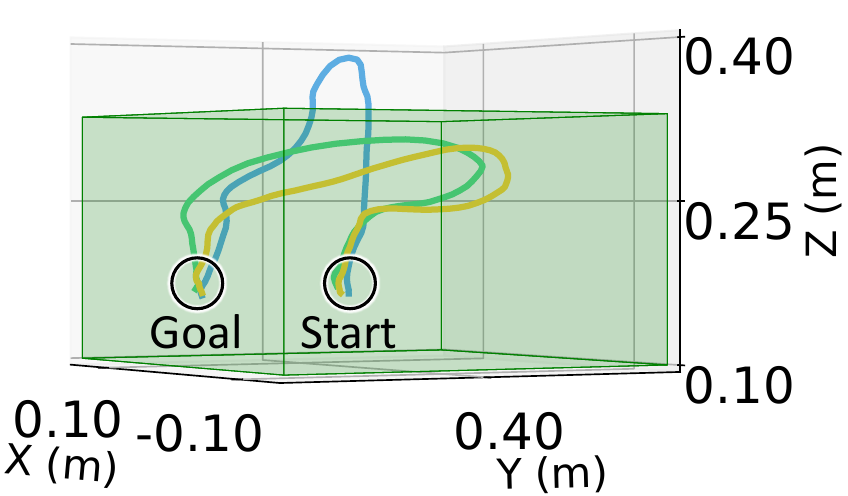}}
        % \vspace{-6mm}
        %\subcaption{Offline + ONline Feedback}
    \end{minipage}
    \begin{minipage}[b]{0.46\linewidth}
        \centering
        \subfloat[\label{fig:experiment:no_trajectory}No FB demo \label{fig:d}]{
        \includegraphics[width=1\hsize]{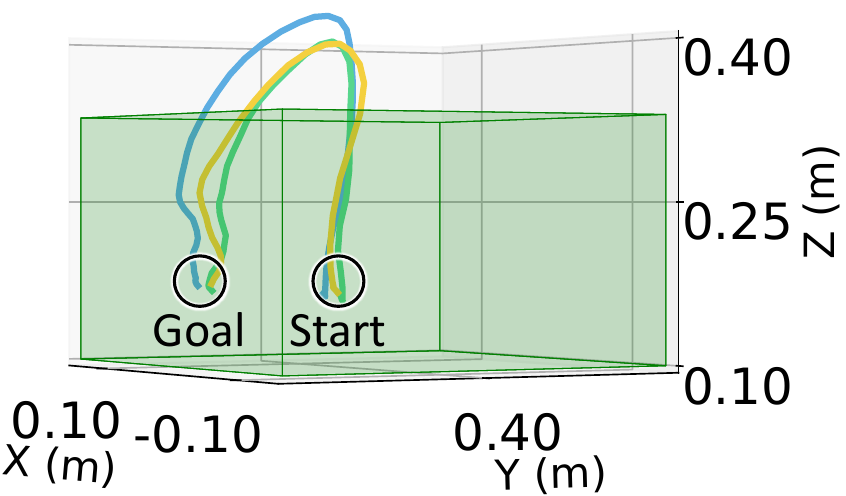}}
        % \vspace{-6mm}
        % \subcaption{No Feedback}
    \end{minipage}
        \caption{
        Transitions of demonstrated motions collected under each demonstration method in peg-insertion task.
        Demonstrated motions were collected using (a) visuo-haptic feedback demonstration and (b) No FB demonstration.
        }
    \label{fig:experiment:pipette_insertion_teaching_trajectory}
\end{figure}
\begin{figure}[t]
    \centering
    \includegraphics[width=0.9\hsize]{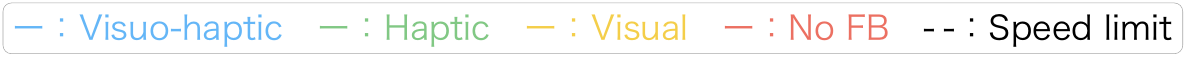}
    % \vspace{-1mm}
    \begin{minipage}{0.7\linewidth}
        \centering
        \includegraphics[width=\hsize]{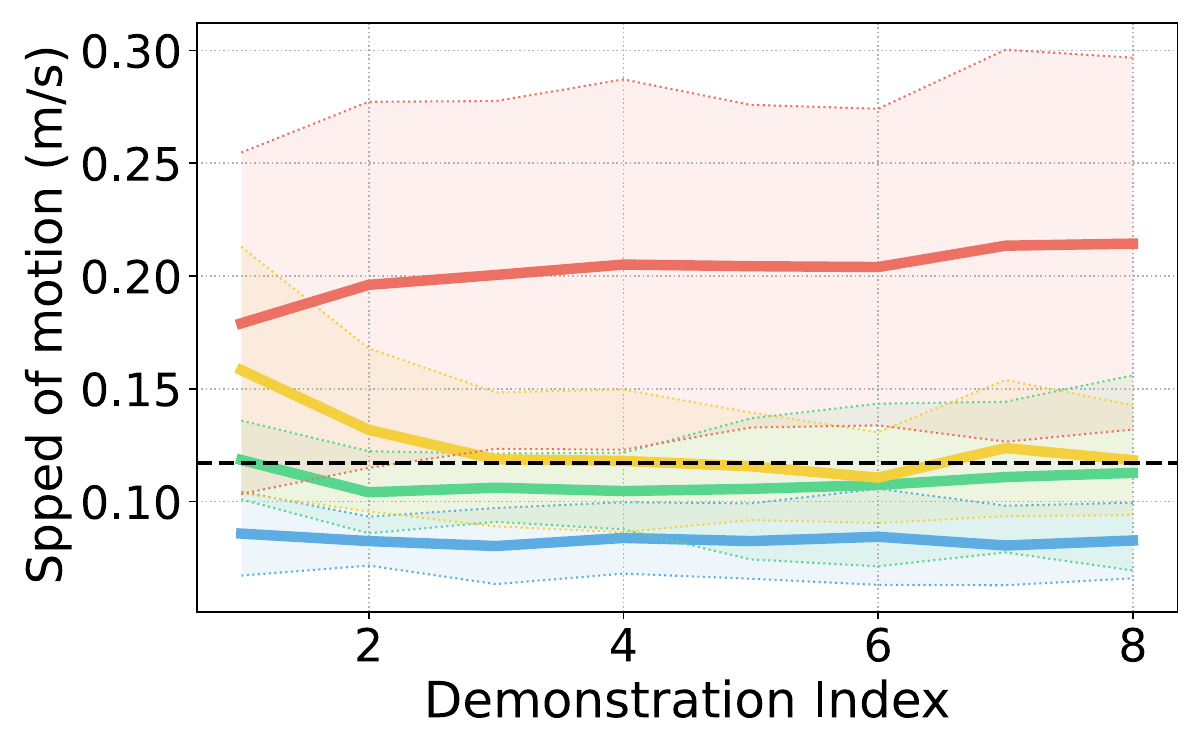}
    \end{minipage}
    \caption{
    Transitions of demonstrated motions collected under each demonstration method in circle-tracing task.
   Horizontal axis represents number of demonstrated motions, and vertical axis represents mean speed of demonstrated motions.
    Each plot shows mean and standard deviation of the demonstrated motion's speed.
    For No FB, statistics are computed across a single experiment by 15 subjects. For the other methods, statistics are computed across a single experiment by five subjects each.
    % \textcolor{red}{For each demonstration method, we first compute the mean speed of each demonstrated motion for each subject, and then report the across-subject mean and standard deviation of these per-subject mean speeds. The results are computed from demonstrations provided by 15 subjects, all of whom performed the No FB demonstrations and were additionally assigned to one of three feedback groups (Visual, Haptic, or Visuo-haptic), with five subjects per feedback group.}
    }
    \label{fig:experiment:circle_tracing_speed}
\end{figure}

For RQ1, we analyzed the feasibility of demonstrated motions in the peg-insertion and circle-tracing tasks.
Fig.~\ref{fig:experiment:feasibility} shows the feasibility transitions for each task.
In both tasks, feasibility is low in the No FB demonstration, whereas feasibility feedback improves the feasibility of the demonstrated motions.

%Fig.~\ref{fig:experiment:pipette_insertion_teaching_trajectory} shows the demonstrated motions change in the peg insertion task, and Fig.~\ref{fig:experiment:circle_tracing_speed} shows the changes in demonstrated speed in the circle tracing task.
% In the peg insertion task, No FB demonstrations often deviated the workspace, while visuo-haptic feedback demonstrations converge to motions that remain within the workspace as demonstrations are repeated.
% We also observe that visual and haptic feedback correct demonstrations to stay within the workspace.
% In the circle tracing task, No FB demonstrations exceed the robot's speed limit, whereas demonstrations with feedback remain within the limit.
% In particular, methods that include haptic feedback strongly suppress overspeed motions, suggesting that haptic feedback helps demonstrators adjust their motions during demonstration.
Fig.~\ref{fig:experiment:pipette_insertion_teaching_trajectory} shows the demonstrated motion's change in the peg-insertion task, and Fig.~\ref{fig:experiment:circle_tracing_speed} shows the changes in demonstrated speed in the circle-tracing task.
% In the peg insertion task, No FB demonstration often included motions that moved outside the workspace, whereas visuo-haptic feedback demonstration showed that, as the number of demonstrated motions increased, subjects provided motions that stayed within the workspace.
% We also confirmed that, in both visual feedback demonstration and haptic feedback demonstration, subjects corrected their demonstrated motions to remain within the workspace in response to feasibility feedback.
In the peg-insertion task, the No FB demonstration often included motions that moved outside the workspace, whereas the demonstration with feasibility feedback gradually shifted toward motions that stayed within the workspace as subjects provided more demonstrated motions.
This trend was consistently observed across feedback modalities, indicating that feasibility feedback helped subjects refine their demonstrated motions to remain within the workspace.
In the circle-tracing task, demonstrated motions under the No FB demonstration often exceeded the robot's end-effector speed limit, whereas the demonstrated motions with feedback stayed within the speed limit.
In particular, demonstration methods that include haptic feedback markedly suppressed overspeed motions, likely because the online haptic feedback provided during motion performance enabled subjects to adjust their motion generation immediately.

% Overall, feasibility feedback improves demonstration feasibility in both tasks.
% It promotes convergence to workspace-feasible motions in the peg insertion task and to speed-feasible motions in the circle tracing task, indicating that FABCO's feasibility feedback is effective across tasks with different control characteristics.
Overall, feasibility feedback improved demonstrated motions' feasibility in both tasks.
It promotes demonstrations that increasingly remain within the workspace in the peg-insertion task and within the speed limit in the circle-tracing task, indicating that FABCO's feasibility feedback is effective across tasks with different control characteristics.

%%%%%%%%%%%%%%%%%%%%%%%%%%%%%%%%%%%%%%%%%%%%%%%%%%%%%%%%%%
\subsubsection{Comparison of workloads and preferences of subjects for each feedback method}
\label{experiment:preference}

\begin{figure}[!t]
    \centering
    \includegraphics[width=0.46\hsize]
    {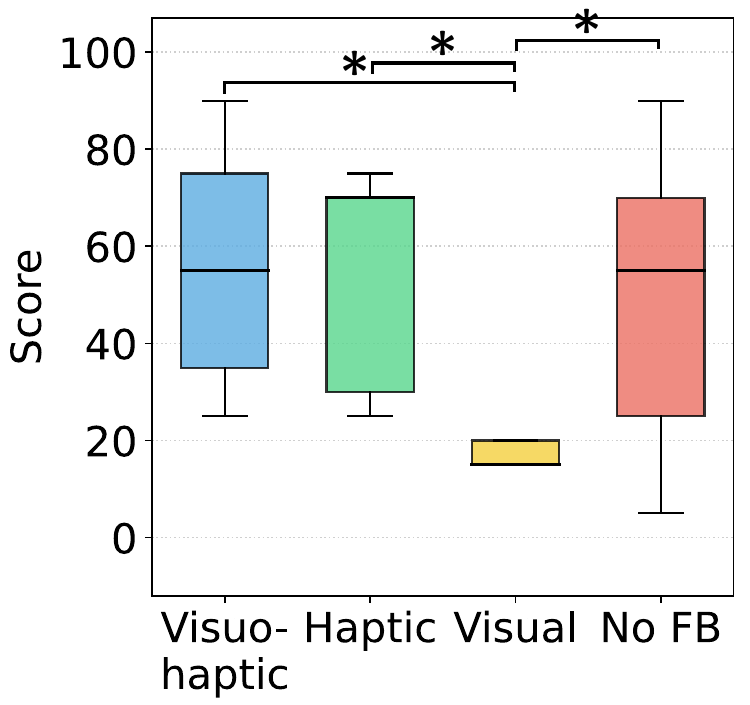}
    \caption{
    NASA-TLX performance subscale in peg-insertion task for all demonstration methods.
    ``*" indicates a statistically significant difference with $p < 0.05$ in t-test.
    }
    \label{fig:experiment:tlx_pipette_insertion}
\end{figure}

\begin{figure}[t]
    \centering
    \begin{minipage}[b]{0.46\linewidth}
        \centering
        \subfloat[Effort]{
            \includegraphics[width=\linewidth]{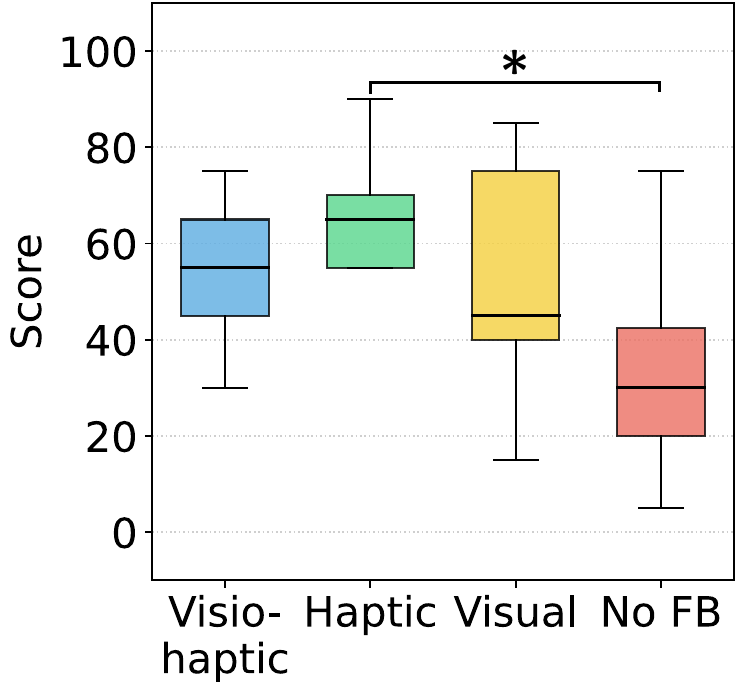}
        }
    \end{minipage}
    \begin{minipage}[b]{0.46\linewidth}
        \centering
        \subfloat[Frustration]{
            \includegraphics[width=\linewidth]{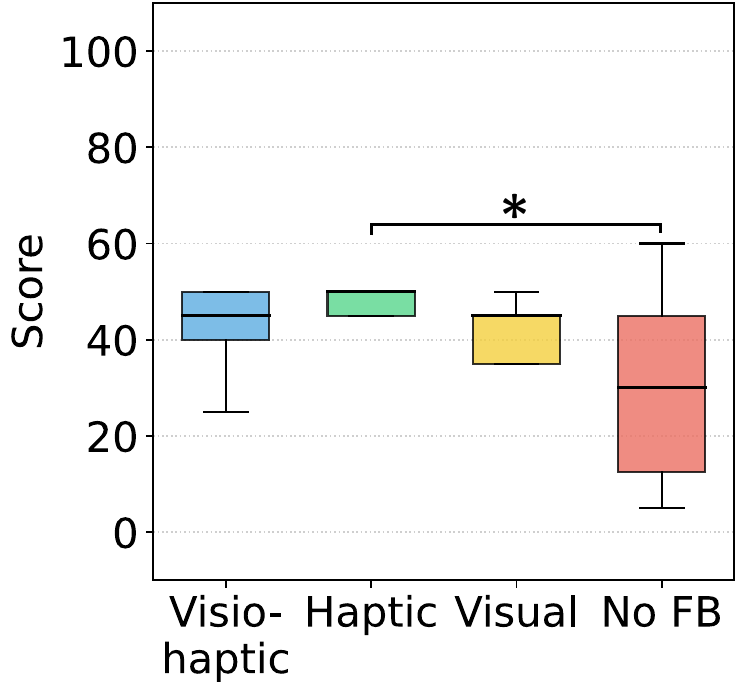}
        }
    \end{minipage}
    \caption{
    Results of NASA-TLX for all demonstration methods in circle-tracing task:
    (a) effort and (b) frustration subscales.
    ``*" indicates a statistically significant difference with $p < 0.05$ in t-test.
    }
    \label{fig:experiment:tlx_circle_trace}
\end{figure}

\begin{figure}[t]
    \centering
    \begin{minipage}[b]{0.46\linewidth}
        \subfloat[Overall vs. pose-level \label{fig:experiment:pose-level_vs_overall}]{\includegraphics[width=\hsize]{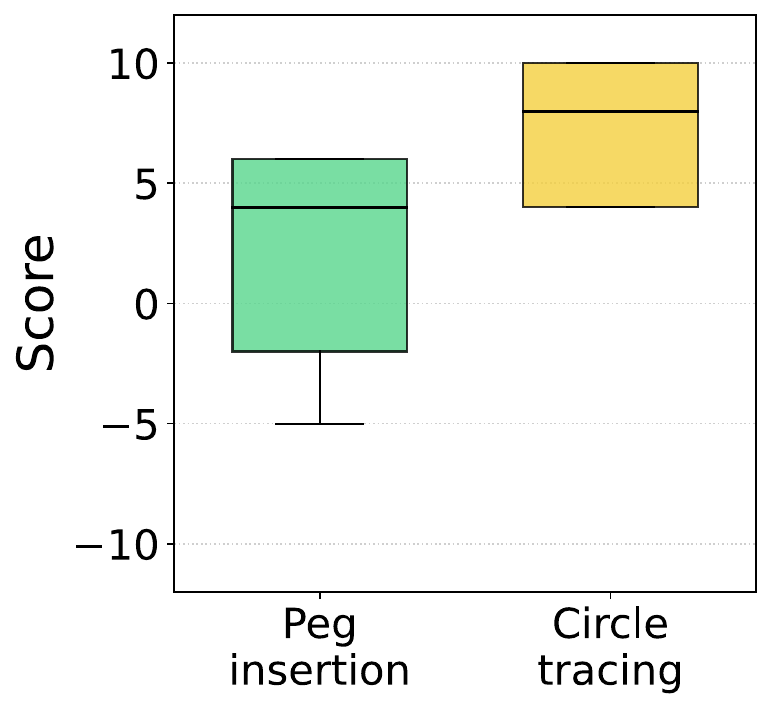}}
    \end{minipage}
    \begin{minipage}[b]{0.46\linewidth}
        \centering
        \subfloat[Visual vs. haptic \label{fig:experiment:haptic_vs_visual}]{\includegraphics[width=\hsize]{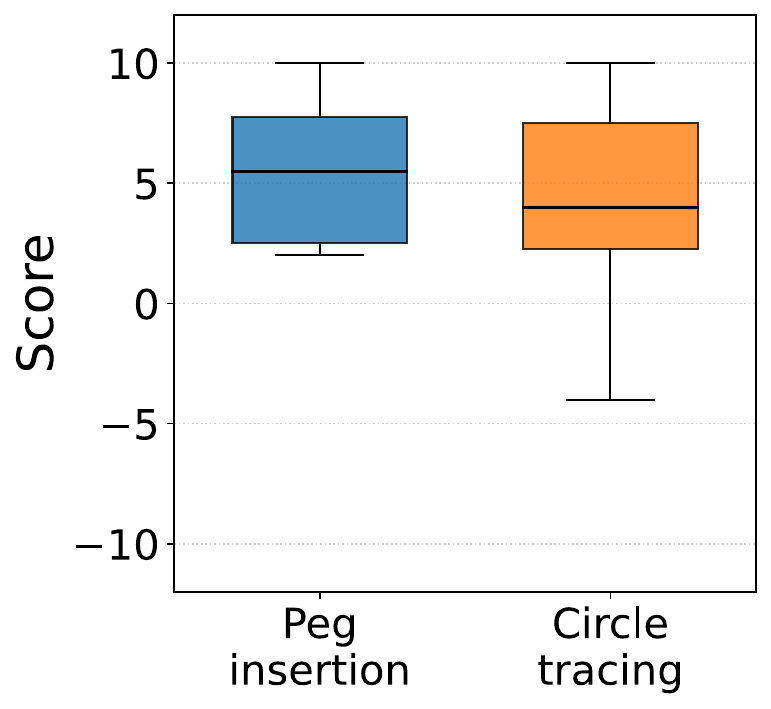}}
    \end{minipage}
    \caption{
    Results of questionnaire on subjects' preferences for demonstration methods.
    (a) Questionnaire results evaluating whether overall visualization or pose-level visualization was more beneficial for subjects using visuo-haptic and visual feedback demonstrations. 
    Lower values indicate a stronger preference for pose-level visualization, while higher values indicate a stronger preference for overall visualization.
    (b) Questionnaire results evaluating whether visual feedback or haptic feedback was more beneficial for subjects using a visuo-haptic feedback demonstration.
    Lower values indicate a stronger preference for visual feedback, while higher values indicate a stronger preference for haptic feedback.
    In both questionnaires, responses were rated on a 20-point scale.
    }
\end{figure}

% For RQ2, we investigate the workload induced by feasibility feedback for the 15 subjects in the peg insertion and circle tracing tasks.
% Among the six NASA-TLX subscales, Fig.~\ref{fig:experiment:tlx_pipette_insertion} shows the results for the peg insertion task that exhibit the most notable differences, and Fig.~\ref{fig:experiment:tlx_circle_trace} shows those for the circle tracing task.
For RQ2, we investigated the workload induced by feasibility feedback for the 15 subjects in the peg-insertion and circle-tracing tasks.
Among the six NASA-TLX subscales, we presented \textit{Performance} for the peg-insertion task (Fig.~\ref{fig:experiment:tlx_pipette_insertion}) and \textit{Effort} and \textit{Frustration} for the circle-tracing task (Fig.~\ref{fig:experiment:tlx_circle_trace}), since these subscales showed the most notable differences among the demonstration methods.
In the peg-insertion task, the performance subscale indicates that visual feedback demonstration yields a significantly lower workload than the other methods.
This is likely because visual feedback presents feasibility after each demonstrated motion, making it easier to review and improve motions, which is particularly helpful for a task requiring high positioning accuracy.
In contrast, in the circle-tracing task, haptic feedback demonstration yields a significantly higher workload in the Effort and Frustration subscales.
Because haptic feedback provides online feasibility information, subjects must continuously check whether their motions satisfy the robot's constraints, which likely increases workload in a task that requires continuous speed control.
Although the visuo-haptic feedback demonstration also includes haptic feedback, its workload is lower than that of the haptic feedback demonstration, suggesting that visual feedback helps reduce cognitive demand during motion performance.

Fig.~\ref{fig:experiment:pose-level_vs_overall} shows preferences between pose-level and overall feasibility visualization in visual feedback.
In both tasks, subjects tended to prefer pose-level visualization.
Since feasibility is presented after each demonstrated motion is performed, subjects have time to review the pose-level visualization in detail, and thus the richer visualization is perceived as beneficial.

% Fig.~\ref{fig:experiment:haptic_vs_visual} shows preferences between visual and haptic feedback.
% In both tasks, subjects tended to prefer haptic feedback.
% In the peg insertion task, some subjects preferred visual feedback, whereas in the circle tracing task haptic feedback was highly valued because it enables online adjustment during demonstration, matching the task characteristics.

% Overall, these results indicate that the feedback modality and timing affect demonstrators' workload and subjective evaluations, and the effect depends on task control characteristics.
% They also suggest that the perceived value of feedback granularity may relate to when feedback is presented.
% Selecting feedback methods according to task characteristics may therefore improve demonstrations without excessively increasing demonstrators' workload.

Fig.~\ref{fig:experiment:haptic_vs_visual} shows the questionnaire results on subjects' preferences between visual and haptic feedback.
In both tasks, subjects tended to prefer haptic feedback over visual feedback.
In the peg-insertion task, however, we also observed a subset of subjects who preferred visual feedback.
This may be because, for tasks requiring high positioning accuracy, some subjects found it easier to understand their performance by visualizing the demonstrated motion.
In contrast, in the circle-tracing task, where speed control is particularly important, the haptic feedback was highly valued because it allowed online adjustment during motion performance, suggesting that haptic feedback is well suited to the task characteristics.

These questionnaire results indicate that the choice of demonstration method affects subjects' workload and preferences, and that the effect depends on the task characteristics.
Specifically, a demonstration method that includes haptic feedback was preferred by many subjects, especially when continuous speed control was important, but it also tended to increase workload.
In contrast, a demonstration method that includes visual feedback was preferred by some subjects when high positioning accuracy was required, and it was also associated with a stronger sense of accomplishment.
These results suggest that selecting the demonstration method based on task characteristics can support an imitation learning framework that improves demonstrated motions without excessively increasing subjects' workload.

% These questionnaire results indicate that the choice of demonstration method affects subjects' workload and preferences, and that the effect depends on the task characteristics.
% Specifically, haptic feedback system was preferred by many subjects, especially when continuous speed control was important, but it also tended to increase workload.
% In contrast, visual feedback system was preferred by some subjects when high positioning accuracy was required, and it was also associated with a stronger sense of accomplishment.
% These results suggest that selecting the demonstration method according to task characteristics can support an imitation learning framework that improves demonstrated motions without excessively increasing subjects' workload.

%%%%%%%%%%%%%%%%%%%%%%%%%%%%%%%%%%%%%%%%%%%%%%%%%%%%%%%%%%%%%%%%%%%%%%%%%%%%%%%%
\subsubsection{Influence of feasibility feedback on learned policy performance}
\label{experiment:success_rate}

\begin{figure}[!t]
    \centering
    \begin{minipage}[b]{0.46\linewidth}
        \centering
        \subfloat[Peg-insertion task \label{fig:experiment:robot:motion:peg_insertion}]{
            \includegraphics[width=\hsize]{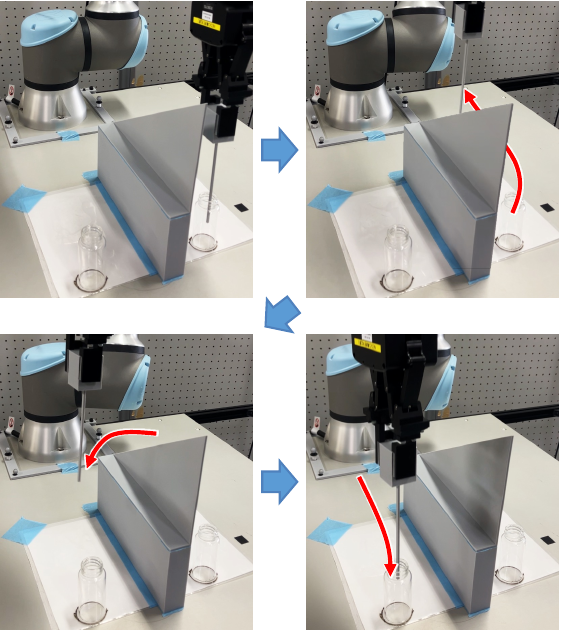}
        }
    \end{minipage}
    \begin{minipage}[b]{0.46\linewidth}
        \centering
        \subfloat[Circle-tracing task\label{fig:experiment:robot:motion:circle_trace}]{
            \includegraphics[width=\hsize]{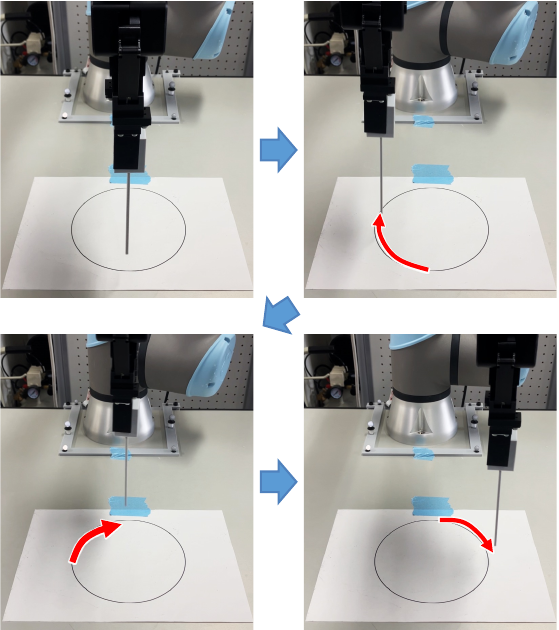}
        }
    \end{minipage}
    \caption{Robot motions using the policy learned by FABCO with visuo-haptic feedback demonstration in (a) peg-insertion task and (b) circle-tracing task.}
    \label{fig:experiment:robot:motion}
\end{figure}

For RQ3, we evaluated task success rates of policies learned from data collected under each demonstration method.
Fig.~\ref{fig:experiment:robot:motion} shows robot motions when executing the peg-insertion task and the circle-tracing task with the policy learned by FABCO from demonstration data collected by the visuo-haptic feedback demonstration method.
For both tasks, we confirmed that the learned policies could execute successful motions that complete the task.

\begin{table}[t]
    \centering
    \caption{
    Pairwise significance comparison of success rates among demonstration methods in peg-insertion task.
    Pairwise t-test results are shown for FABCO, with visuo-haptic, haptic, and visual feedback, and for BCO.
    ``+'' indicates that the method in the row achieves a higher success rate than the method in the column, while ``–'' indicates the opposite.
    ``*'' denotes statistical significance ($p < 0.05$).
   Bottom row shows the success rate (\%) for each method.
    }
    \label{table:experiment:success:rate:peg_insertion}
    \begin{tabular}{l|cccc}
        \toprule
                     & FABCO with   & FABCO       & FABCO   & \\
                     & visuo-haptic & with haptic & with visual      & BCO \\
        \midrule       
        \makecell[l]{FABCO with \\ visuo-haptic} 
                     &\diagbox{}{}  & +$*$        & +$*$        & +$*$        \\
        \makecell[l]{FABCO \\ with haptic} 
                     & -$*$         &\diagbox{}{} & -$*$        & +$*$        \\
        \makecell[l]{FABCO \\ with visual} 
                     & -$*$         & +$*$        &\diagbox{}{} & +$*$        \\
        BCO          & -$*$         & -$*$        & -$*$        &\diagbox{}{} \\
        \midrule       
        Success rate & 90.0         & 66.0        & 78.0        & 0.0 \\
        \bottomrule
    \end{tabular}
\end{table}
\begin{figure}[t!]
    \centering
    \includegraphics[width=0.75\hsize]{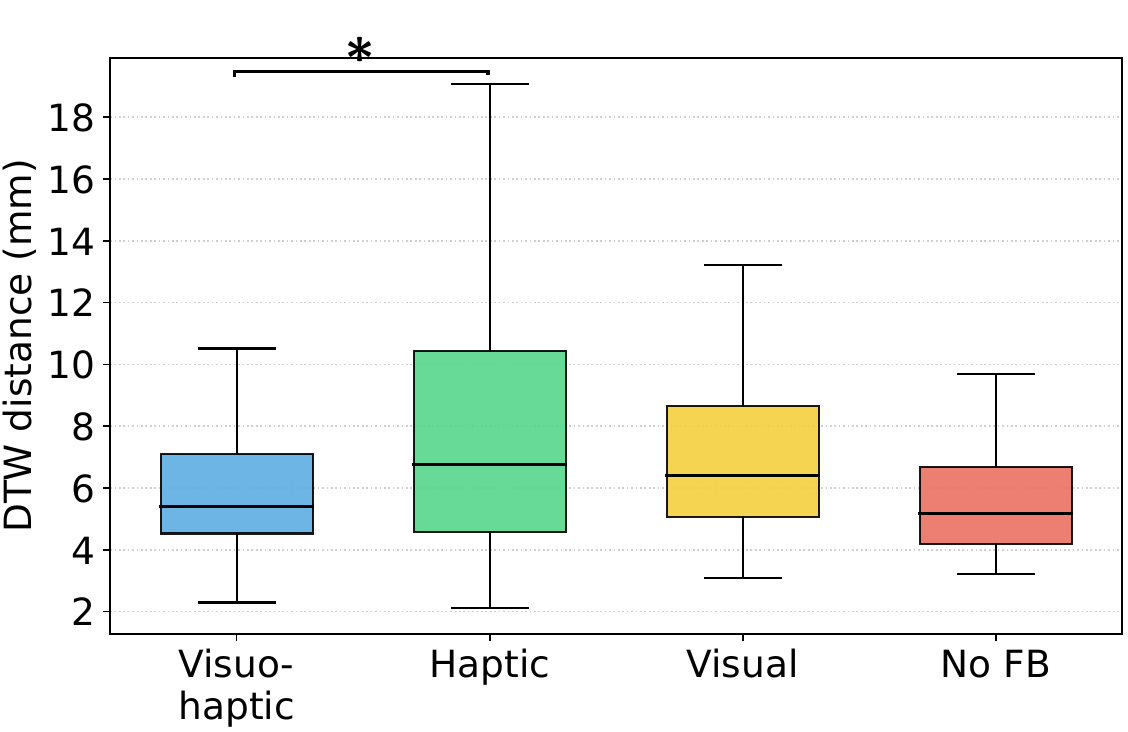}
    \vspace{-3mm}
    \caption{
    Dynamic Time Warping (DTW) distances between mean demonstrated motion and each demonstrated motion for each subject in peg-insertion task.
   Mean demonstrated motion is computed by averaging positions at each time step across subjects' temporally normalized demonstrated motions.
    Statistical significance of DTW differences among  methods is evaluated using pairwise t-tests across all combinations of demonstration methods, where ``*'' indicates a statistically significant difference ($p < 0.05$).
    }
    \label{fig:experiment:dtw}
\end{figure}

Table~\ref{table:experiment:success:rate:peg_insertion} presents success rates of the peg-insertion task and Wald test results.
Success rates are highest for FABCO with visuo-haptic feedback demonstration, followed by visual feedback demonstration, haptic feedback demonstration, and BCO, with statistically significant differences for all pairs of demonstration methods.
BCO learned on No FB demonstration data achieves the lowest success rate, consistent with the demonstration feasibility results shown in Fig.~\ref{fig:experiment:pipette_insertion_feasibility}.
However, although the feasibility results indicate that  data collected by haptic feedback demonstration exhibit higher feasibility than those collected with visuo-haptic and visual feedback demonstration, the learned policy success rate is lower than those of the visuo-haptic and visual feedback demonstrations.
To clarify this discrepancy, Fig.~\ref{fig:experiment:dtw} analyzes demonstrated motion variability across demonstration methods.
Haptic feedback demonstration results in greater variability in demonstrated motions, which likely degrades policy performance.
In contrast, visuo-haptic feedback demonstration and visual feedback demonstration increase feasibility while also facilitating more consistent motion refinement, contributing to improved policy performance.

Table~\ref{table:experiment:success:rate:circle_tracing} reports success rates of the circle-tracing task and Wald test results.
Success rates are highest for FABCO with visuo-haptic feedback demonstration, followed by haptic feedback demonstration, visual feedback demonstration, and BCO; moreover, statistically significant differences are observed for all pairs except visuo-haptic vs.\ haptic.
% In the circle tracing task, exceeding the end-effector speed limit is the primary factor that reduces feasibility; therefore, haptic feedback demonstration and visuo-haptic feedback demonstration can immediately indicate speed-limit violations to subjects, which efficiently promotes motion refinement and improves policy performance.
In the circle-tracing task, exceeding the end-effector speed limit is the primary factor reducing feasibility; therefore, haptic feedback demonstration and visuo-haptic feedback demonstration can immediately inform subjects when their motion exceeds the speed limit, which efficiently promotes motion refinement and improves policy performance.
Across both tasks, FABCO with visuo-haptic feedback demonstration achieves the highest success rate, indicating that combining the visual and haptic modalities of feedback supports the learning of higher-performing policies.

\begin{table}[t]
    \centering
    \caption{
    Pairwise significance comparison of success rates among demonstration methods in circle-tracing task.
    Pairwise t-test results are shown for FABCO, with visuo-haptic, haptic, and visual feedback, and for BCO.
    ``+'' indicates that the method in the row achieves a higher success rate than the method in the column, while ``–'' indicates the opposite.
    ``*'' denotes statistical significance ($p < 0.05$).
    The bottom row shows the success rate (\%) for each method.
    }
    \label{table:experiment:success:rate:circle_tracing}
    \begin{tabular}{l|cccc}
        \toprule
                     & FABCO with   & FABCO       & FABCO   & \\
                     & visuo-haptic & with haptic & with visual      & BCO \\
        \midrule      
        \makecell[l]{FABCO with \\ visuo-haptic} 
                     &\diagbox{}{}  & n.s.        & +$*$        & +$*$ \\
        \makecell[l]{FABCO \\ with haptic} 
                     & n.s.         &\diagbox{}{} & +$*$        & +$*$ \\
        \makecell[l]{FABCO \\ with visual} 
                     & -$*$         & -$*$        &\diagbox{}{} & +$*$ \\
        BCO          & -$*$         &  -$*$       & -$*$        & \diagbox{}{} \\
        \midrule
        Success rate & 88.0         & 80.0        & 60.0        & 26.7 \\
        \bottomrule
    \end{tabular}
\end{table}
\begin{figure}[t]
    \centering
    \begin{minipage}[b]{1\linewidth}
        \centering
        \includegraphics[width=0.4\hsize]{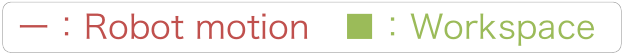}
    \end{minipage}
    % First row
    \begin{minipage}[b]{0.46\linewidth}
        \centering
        \subfloat[FABCO with visuo-haptic demo \label{fig:pipette_insertion:a}]{
        \includegraphics[width=1\hsize]{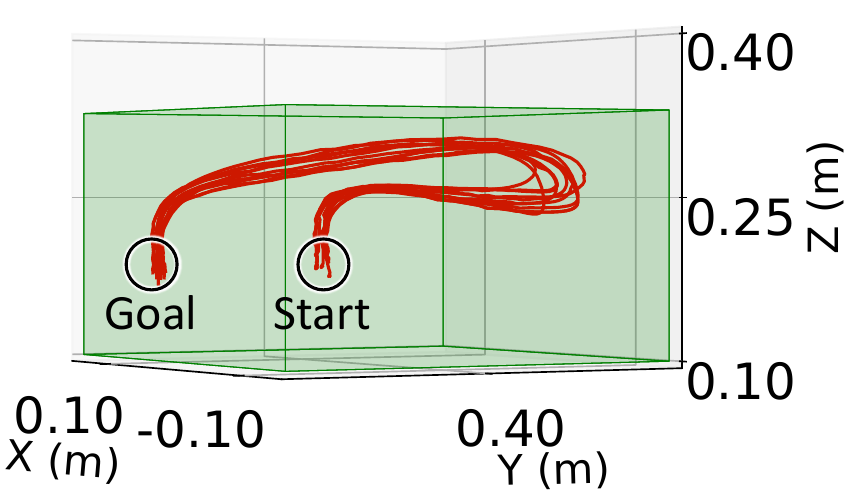}}
        % \vspace{-6mm}
        %\subcaption{Offline + ONline Feedback}
    \end{minipage}
    \begin{minipage}[b]{0.46\linewidth}
        \centering
        \subfloat[FABCO with haptic demo
        \label{fig:pipette_insertion:b}]{
        \includegraphics[width=1\hsize]{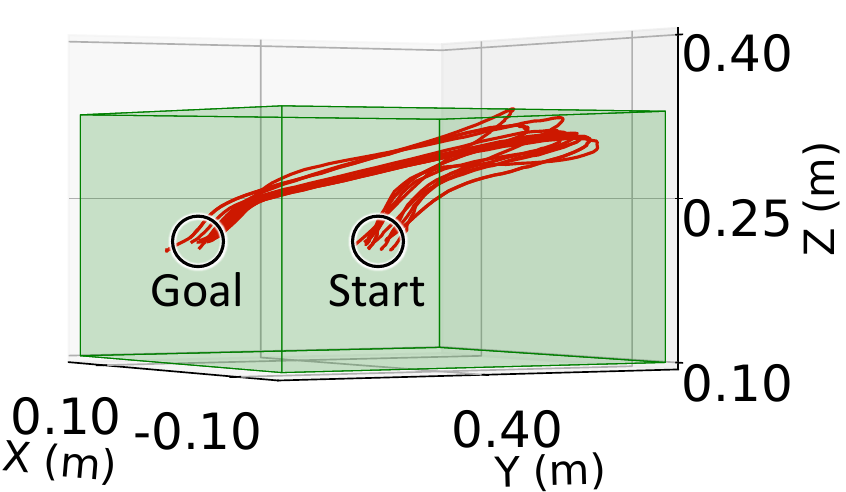}}
        % \vspace{-6mm}
        % \subcaption{Haptic Feedback}
    \end{minipage}
    \begin{minipage}[b]{0.46\linewidth}
        \centering
        \subfloat[FABCO with visual demo \label{fig:pipette_insertion:c}]{
        \includegraphics[width=1\hsize]{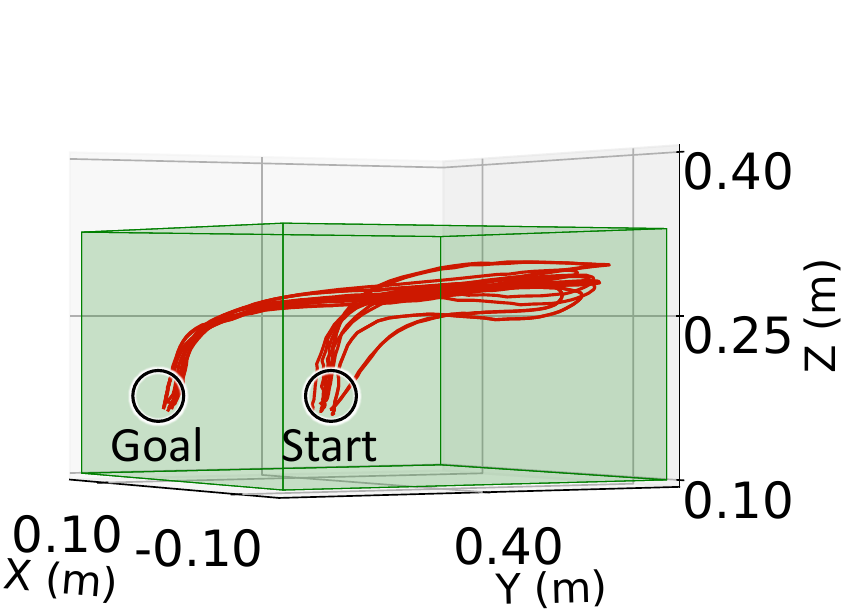}}
        % \vspace{-6mm}
        % \subcaption{Offline Feedback}
    \end{minipage}
    \begin{minipage}[b]{0.46\linewidth}
        \centering
        \subfloat[BCO with No FB demo \label{fig:pipette_insertion:d}]{
        \includegraphics[width=1\hsize]{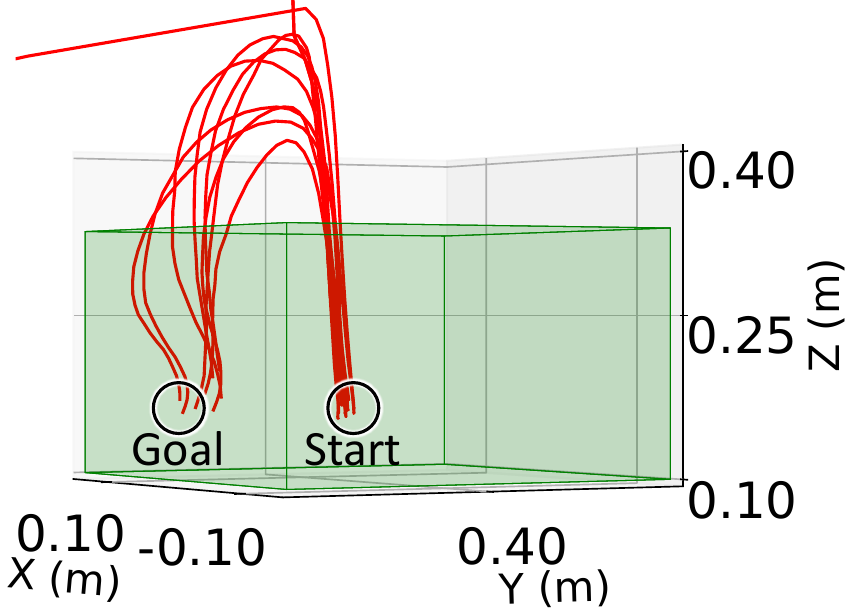}}
        % \vspace{-6mm}
        % \subcaption{No Feedback}
    \end{minipage}
    \caption{
    Robot motion trajectories of policies learned in peg-insertion task.
    Policies learned by 
    (a) FABCO with visuo-haptic demonstration,
    (b) FABCO with haptic demonstration,
    (c) FABCO with visual demonstration, and
    (d) BCO with No FB demonstration.
   Red lines and green areas indicate robot motion trajectories and the workspace, respectively.
    }
    \label{fig:experiment:pipette_insertion_policy_trajectory}
\end{figure}

\begin{figure}[!t]
    \centering
    \begin{minipage}[b]{1\linewidth}
        \centering
        \includegraphics[width=0.43\hsize]{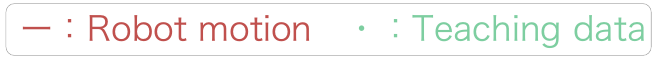} \\
    \end{minipage}
    % First row
    \begin{minipage}[b]{0.46\linewidth}
        \centering
        \subfloat[FABCO with visuo-haptic demo \label{fig:cricle_trace:a}]{
        \includegraphics[width=1\hsize]{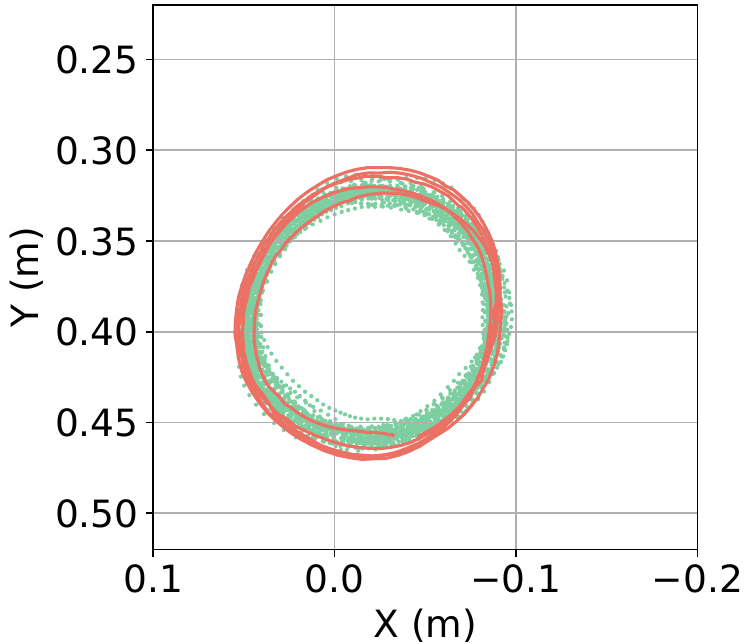}}
    \end{minipage}
    \begin{minipage}[b]{0.46\linewidth}
        \centering
        \subfloat[FABCO with haptic demo \label{fig:cricle_trace:b}]{
        \includegraphics[width=1\hsize]{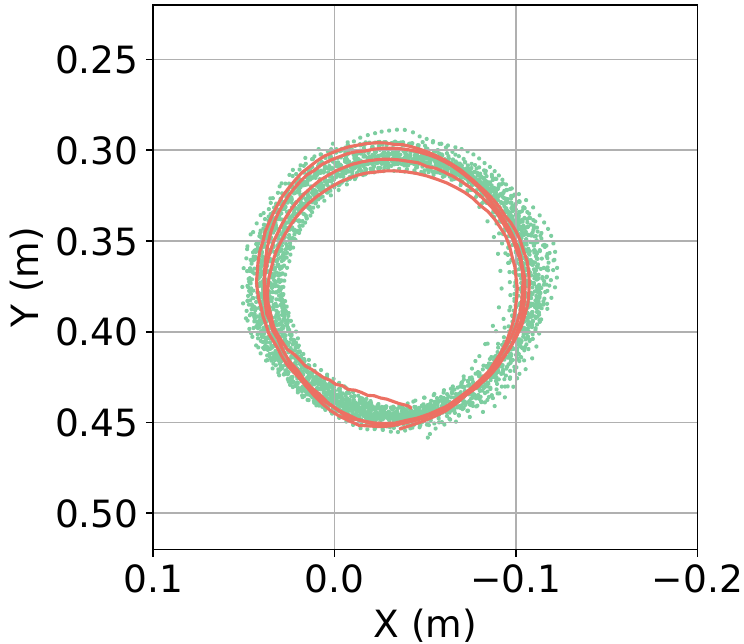}}
    \end{minipage}
    \begin{minipage}[b]{0.46\linewidth}
        \centering
        \subfloat[FABCO with visual demo \label{fig:cricle_trace:c}]{
        \includegraphics[width=1\hsize]{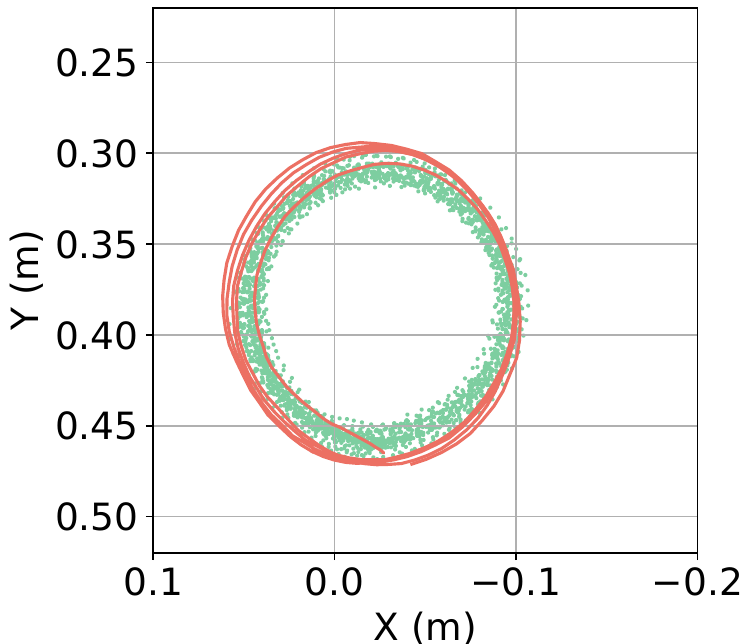}}
    \end{minipage}
    \begin{minipage}[b]{0.46\linewidth}
        \centering
        \subfloat[BCO with No FB demo \label{fig:cricle_trace:d}]{
        \includegraphics[width=1\hsize]{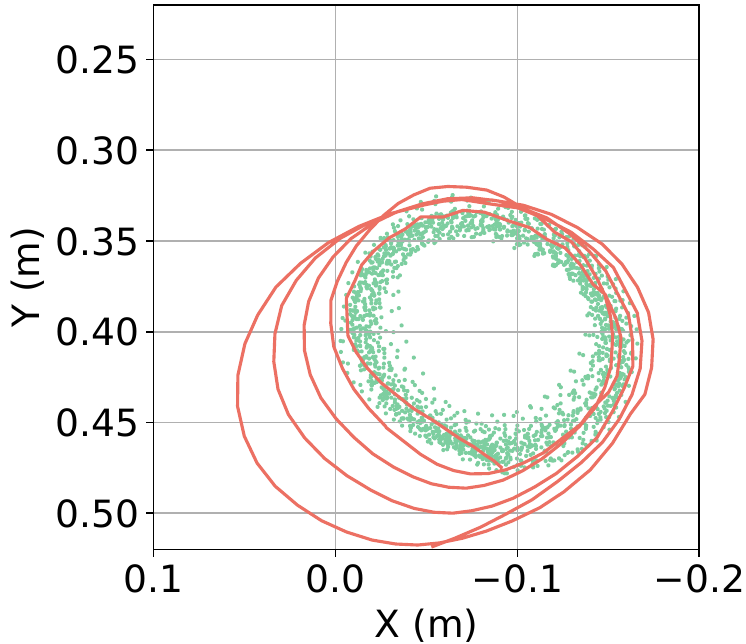}}
    \end{minipage}
    \caption{
    Robot motions of learned policies in circle-tracing task.
   Policies learned by 
    (a) FABCO with visuo-haptic demonstration,
    (b) FABCO with haptic demonstration,
    (c) FABCO with visual demonstration, and
    (d) BCO with No FB demonstration.
   Red lines and green points indicate robot motion and demonstrated motion, respectively.
    }
    \label{fig:experiment:circle_trace_policy_trajectory}
\end{figure}

Fig.~\ref{fig:experiment:pipette_insertion_policy_trajectory} and Fig.~\ref{fig:experiment:circle_trace_policy_trajectory} show robot motions generated by the learned policies for peg insertion and circle tracing, respectively.
In both tasks, policies learned on data collected with feedback demonstration execute motions stably without large deviations from the demonstrated motions.
In contrast, the policy learned by BCO learns motions that gradually drift away from the demonstrated motion over time.
This is likely because the No FB demonstration used for BCO includes many low-feasibility motions, which makes accurate action estimation difficult and leads to covariate shift, causing the policy to progressively deviate from the demonstrated motions.

Overall, feasibility feedback improves policy performance.
Across both tasks, FABCO with visuo-haptic feedback demonstration achieved the highest success rates, indicating that combining visual and haptic feedback enables the collection of demonstration data that is more suitable for policy learning.
Moreover, comparing feedback modalities suggests complementary roles: haptic feedback is effective for on-the-spot motion refinement, whereas visual feedback is effective for improving the overall quality of demonstrated motions.

%%%%%%%%%%%%%%%%%%%%%%%%%%%%%%%%%%%%%%%%%%%%%%%%%%%%%%%%%%
\subsubsection{Ablation study on feasibility-aware policy learning}
\label{experiment:filtering}

\begin{table}
    \centering
    % \caption{ペグ挿入タスクとまるなぞりタスクでのデータフィルタリングの有無による成功率の比較: 各被験者データから学習した方策を用いて1方策につき10試行(合計50試行)から成功率を算出．$\dagger$ and $\ddagger$ indicate significant differences between FABCO and FABCO w/o Filtering ($p<0.05$)}
    % \caption{
    % Comparison of success rates for the policy learned from FABCO using visuo-haptic demonstration, with and without feasibility-aware policy learning, in the peg insertion and circle tracing tasks.
    % ``$\dagger$" and ``$\ddagger$" indicate a statistically significant difference with $p < 0.05$ in the pairwise t-test.
    % }
    \caption{
    Comparison of success rates for policies learned by FABCO from visuo-haptic demonstrations, with and without feasibility-aware policy learning (FPL), in peg-insertion and circle-tracing tasks.
    Variant learned without FPL denoted as FABCO w/o FPL.
    ``$\dagger$" and ``$\ddagger$" indicate a statistically significant difference with $p < 0.05$ in pairwise t-test.
    }
    \label{table:experiment:success_rate_filtering}
    \begin{tabular}{r|cc|cc}
        \toprule
        & \multicolumn{2}{c|}{Peg-insertion task} & \multicolumn{2}{c}{Circle-tracing task} \\
        \cmidrule(lr){2-3}\cmidrule(lr){4-5}
        &
        \makecell[c]{FABCO} &
        \makecell[c]{FABCO\\w/o FPL} &
        \makecell[c]{FABCO} &
        \makecell[c]{FABCO\\w/o FPL} \\
        \midrule
        First-half & 76.0$^{\dagger}$ & 32.0$^{\dagger}$ & 64.0$^{\ddagger}$ & 36.0$^{\ddagger}$ \\
        Full       & 90.0~            & 88.0~            & 88.0~             & 88.0~ \\
        \bottomrule
    \end{tabular}
\end{table}

For RQ4, we compared success rates and Hausdorff distances between robot executions and the demonstration data for policies learned with and without feasibility-aware policy learning (FPL).
We evaluated two learning conditions: with full data and first-half data, which were collected with the visuo-haptic feedback demonstration.
Table~\ref{table:experiment:success_rate_filtering} summarizes the success rates for each task with and without feasibility-aware policy learning.
The results show that, under the full data condition, both tasks achieve high success rates regardless of whether feasibility-aware policy learning is applied.
In contrast, under the first-half data condition, feasibility-aware policy learning improves success rates for both tasks.

\begin{figure}[t]
    \centering
    % ------- 凡例 -------
    \begin{minipage}[b]{1\linewidth}
        \centering
        \includegraphics[width=0.4\hsize]{fig/robot_motion_trajectory_peg_legend.pdf}
    \end{minipage}

    % ------- 上段：Half data -------
    \begin{minipage}[b]{0.46\linewidth}
        \centering
        \subfloat[FABCO using first-half data \label{fig:pipette_insertion_with_filter_15}]{
        \includegraphics[width=1\hsize]{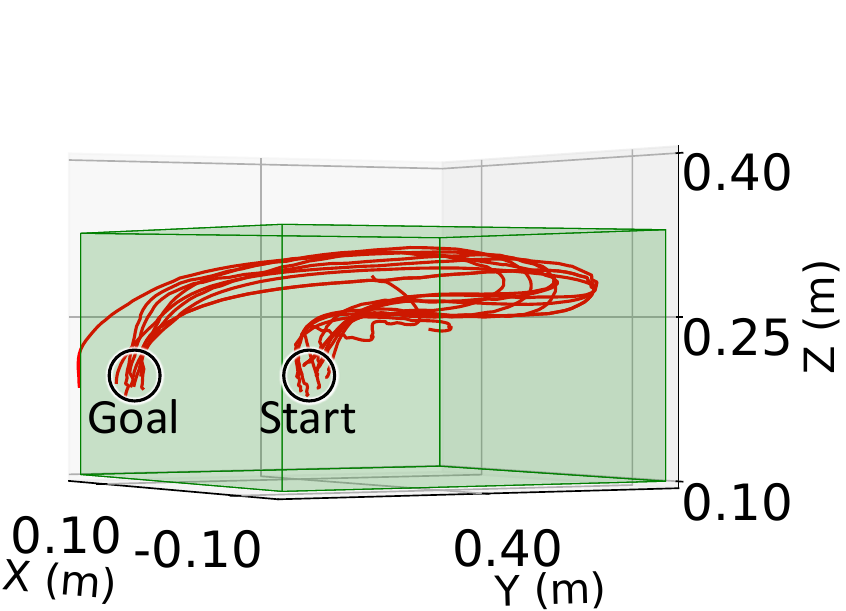}}
    \end{minipage}
    \begin{minipage}[b]{0.46\linewidth}
        \centering
        \subfloat[FABCO w/o FPL using first-half data
        \label{fig:pipette_insertion_without_filter_15}]{
        \includegraphics[width=1\hsize]{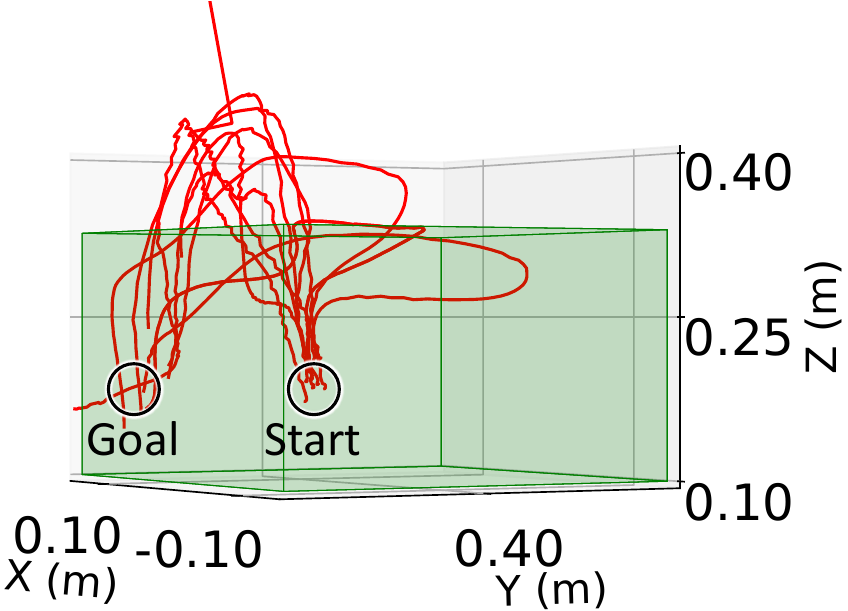}}
    \end{minipage}

    % ------- 下段：Full data -------
    \begin{minipage}[b]{0.46\linewidth}
        \centering
        \subfloat[FABCO using full data \label{fig:pipette_insertion_with_filter_30}]{
        \includegraphics[width=1\hsize]{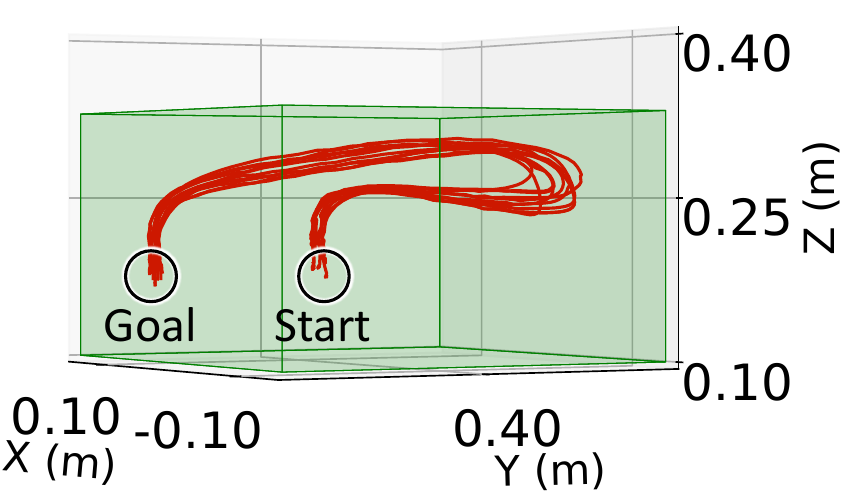}}
    \end{minipage}
    \begin{minipage}[b]{0.46\linewidth}
        \centering
        \subfloat[FABCO w/o FPL using full data \label{fig:pipette_insertion_without_filter_30}]{
        \includegraphics[width=1\hsize]{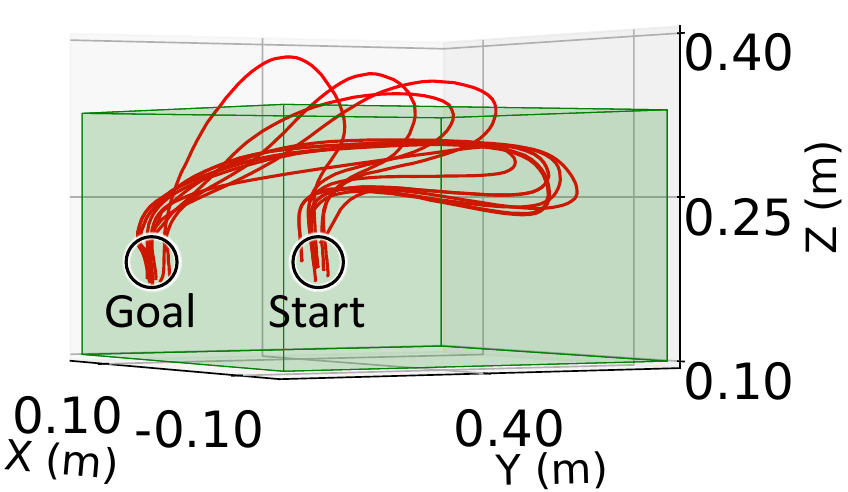}}
    \end{minipage}

    % \caption{Peg insertion trajectories under different numbers of demonstration data.
    % (a,b): Half of demonstration data. (c,d): Full demonstration data.}
    \caption{
    Robot motion of learned policies in peg-insertion task.
    Policies learned by 
    (a) FABCO using first half of demonstration data,
    (b) FABCO w/o feasibility-aware policy learning (FPL) using first half of demonstration data,
    (c) FABCO using full demonstration data, and
    (d) FABCO w/o FPL using full demonstration data.
   Red lines and green areas indicate robot motion trajectories and the workspace, respectively.
    }    \label{fig:experiment:pipette_insertion_policy_trajectory_filter}
\end{figure}
\begin{figure}[t]
    \centering
    % ------- 凡例 -------
    \begin{minipage}[b]{1\linewidth}
        \centering
        \includegraphics[width=0.43\hsize]{fig/robot_motion_circle_legend.pdf} \\
    \end{minipage}

    % ------- 上段：4 data -------
    \begin{minipage}[b]{0.46\linewidth}
        \centering
        \subfloat[FABCO using first-half data \label{fig:circle_trace_with_filter_4}]{
        \includegraphics[width=1\hsize]{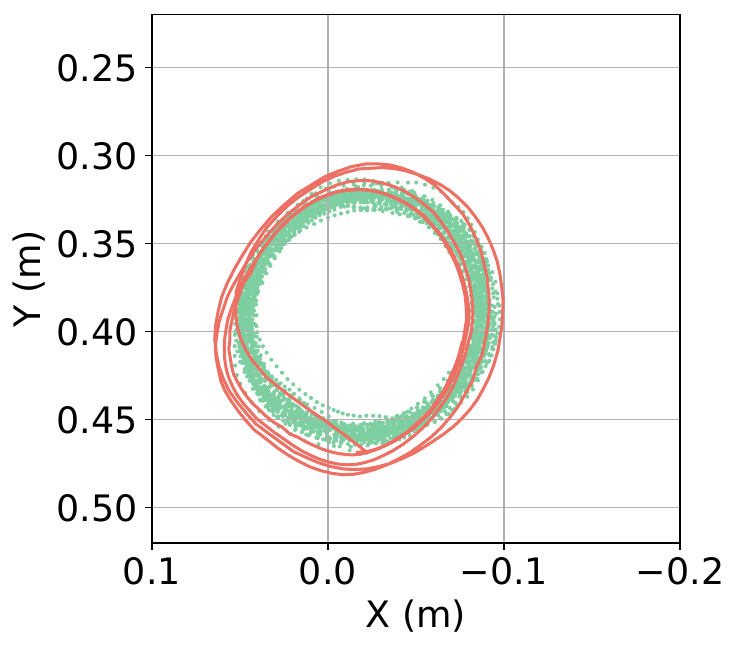}}
    \end{minipage}
    \begin{minipage}[b]{0.46\linewidth}
        \centering
        \subfloat[FABCO w/o FPL using first-half data \label{fig:circle_trace_without_filter_4}]{
        \includegraphics[width=1\hsize]{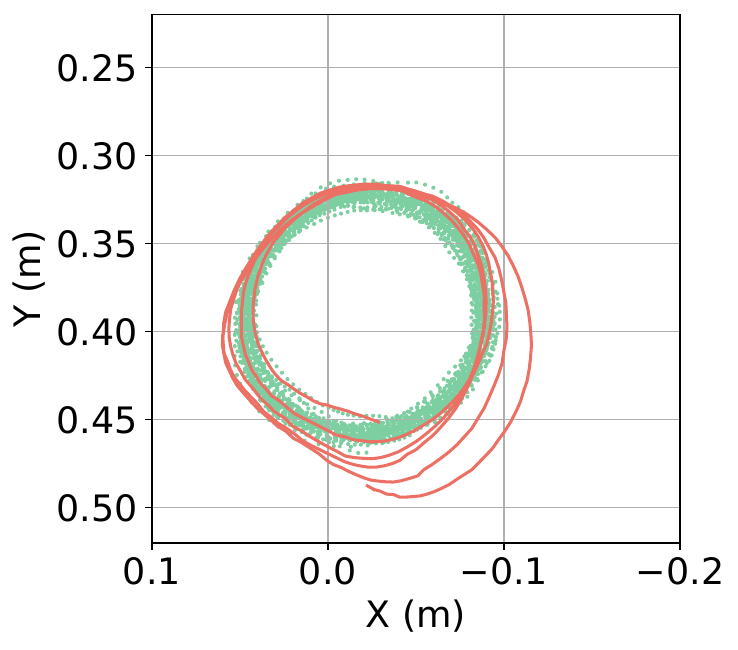}}
    \end{minipage}

    % ------- 下段：8 data -------
    \begin{minipage}[b]{0.46\linewidth}
        \centering
        \subfloat[FABCO using full data \label{fig:circle_trace_with_filter_8}]{
        \includegraphics[width=1\hsize]{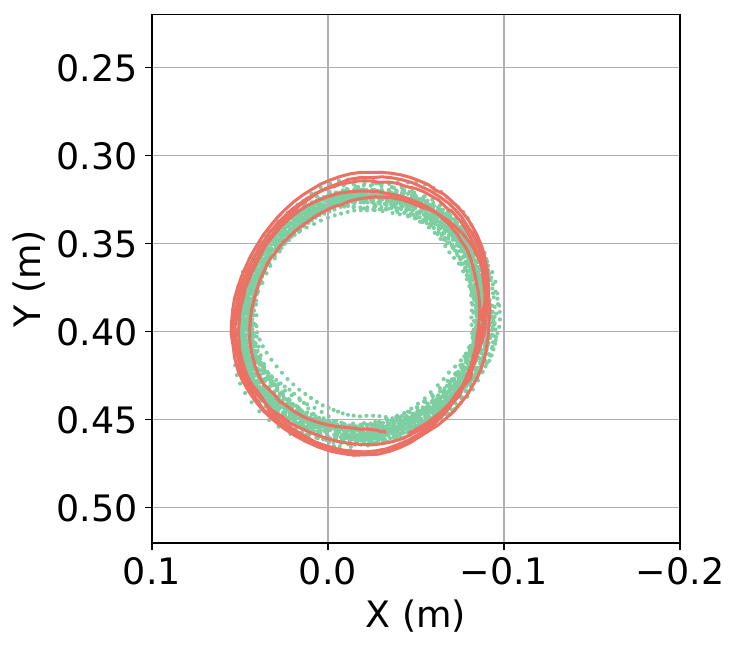}}
    \end{minipage}
    \begin{minipage}[b]{0.46\linewidth}
        \centering
        \subfloat[FABCO w/o FPL using full data \label{fig:circle_trace_without_filter_8}]{
        \includegraphics[width=1\hsize]{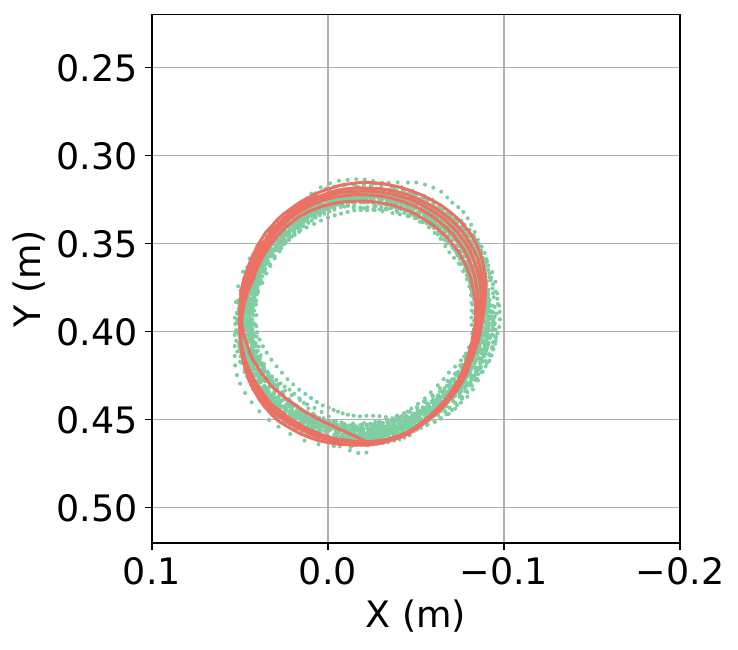}}
    \end{minipage}
    \caption{
    Robot motion trajectories of learned policies in circle-tracing task.
    Policies learned by
    (a) FABCO using first half of demonstration data,
    (b) FABCO w/o feasibility-aware policy learning (FPL) using first half of demonstration data,
    (c) FABCO using full demonstration data, and
    (d) FABCO w/o FPL using full demonstration data.
   Red lines and green points indicate robot motion trajectories and demonstration data, respectively.
    }
\label{fig:experiment:circle_trace_policy_trajectory_filter}
\end{figure}

Fig.~\ref{fig:experiment:pipette_insertion_policy_trajectory_filter} shows the peg-insertion trajectories with and without feasibility-aware policy learning, and Fig.~\ref{fig:experiment:circle_trace_policy_trajectory_filter} shows the corresponding trajectories for the circle-tracing task.
In the peg-insertion task, feasibility-aware policy learning suppresses behaviors that leave the workspace.
In the circle-tracing task, policies learned with full data exhibit little difference, whereas with first-half data, feasibility-aware policy learning reduces deviations from the demonstrated motions.

\begin{figure}[!t]
    \centering
    \begin{minipage}[b]{1\linewidth}
        \centering
        \includegraphics[width=0.5\hsize]{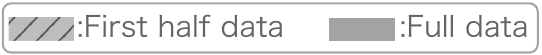}
    \end{minipage}
    \begin{minipage}[b]{0.46\linewidth}
        \subfloat[Peg-insertion task \label{fig:experiment:hausdorff_peg_filtering}]{\includegraphics[width=\hsize]{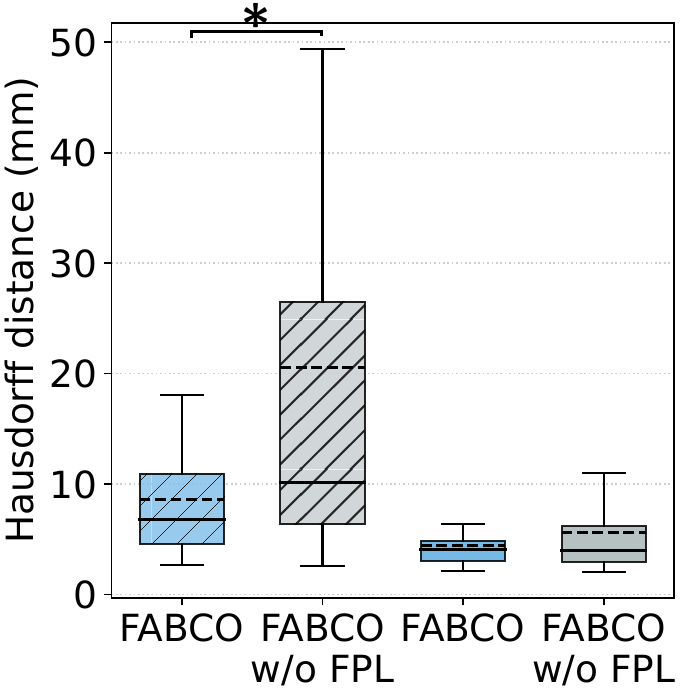}}
    \end{minipage}
    \begin{minipage}[b]{0.46\linewidth}
        \centering
        \subfloat[Circle-tracing task \label{fig:experiment:hausdorff_circle_filtering}]{\includegraphics[width=\hsize]{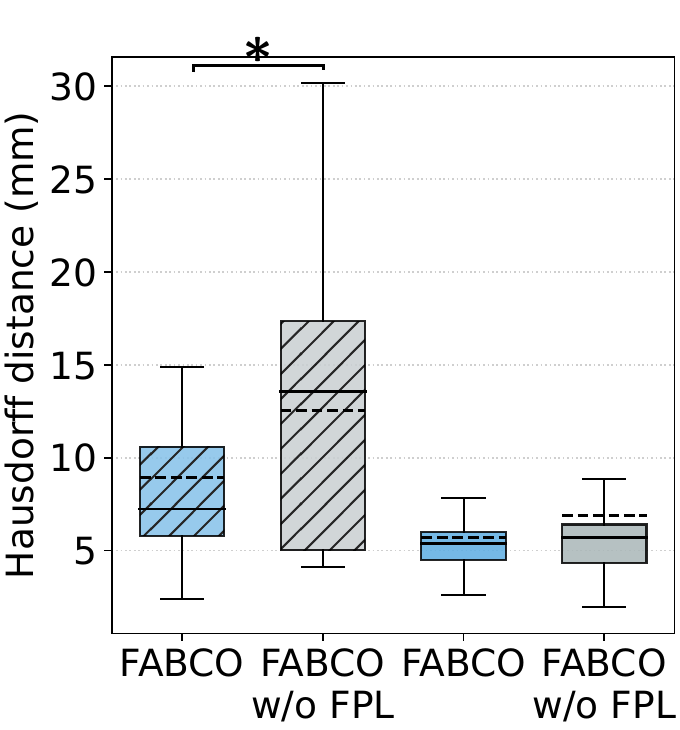}}
    \end{minipage}
    \caption{
    Comparison of Hausdorff distance between demonstrated and robot motions generated by learned policies with and without feasibility-aware policy learning (FPL).
    (a) Results for peg-insertion task.
    (b) Results for circle-tracing task.
    ``*" indicates a statistically significant difference with $p < 0.05$ in pairwise t-test.
    }
    \label{fig:experiment:hausdorff_filtering}
\end{figure}

Fig.~\ref{fig:experiment:hausdorff_filtering} presents the Hausdorff distances between the demonstrated motions and the robot executions for each task.
Under both full and first-half data conditions, feasibility-aware policy learning yields significantly smaller Hausdorff distances, indicating that the learned policies produce motions more similar to the demonstrated motions.
Even in the full data condition, although the success-rate difference is not statistically significant, feasibility-aware policy learning reduces the Hausdorff distance.

These results indicate that feasibility-aware policy learning improves policy stability and performance by emphasizing high-feasibility demonstration data collected through demonstration with feasibility feedback.
In particular, it provides a larger performance gain when the amount of demonstration data are limited.
Moreover, the evaluation of Hausdorff distance suggests that feasibility-aware policy learning suppresses motions that leave the workspace and large deviations from the demonstrated motions, leading to executions that better match the demonstrated motions.

%%%%%%%%%%%%%%%%%%%%%%%%%%%%%%%%%%%%%%%%%%%%%%%%%%%%%%%%%%%%%%%%%%%%%%%%%%%%%%%%
%%%%%%%%%%%%%%%%%%%%%%%%%%%%%%%%%%%%%%%%%%%%%%%%%%%%%%%%%%%%%%%%%%%%%%%%%%%%%%%%
\section{Discussions}

%%%%%%%%%%%%%%%%%%%%%%%%%%%%%%%%%%%%%%%%%%%%%%%%%%%%%%%%%%%%%%%%%%%%%%%%%%%%%%%%
%%%%%%%%%%%%%%%%%%%%%%%%%%%%%%%%%%%%%%%%%%%%%%%%%%%%%%%%%%%%%%%%%%%%%%%%%%%%%%%%

\subsection{Limitations of feedback methods and possible improvements}

In this study, we developed FABCO to estimate the feasibility of demonstrated motions and presented the estimated feasibility to subjects through two feedback modalities to encourage refinement of demonstrated motions.
However, the current approach is limited to indicating which parts of a demonstrated motion are infeasible, and subjects must decide how to modify their motions on their own.
As a result, refining demonstrated motions can depend on subjects' knowledge of robots and the task, and subjects without specialized knowledge may require substantial trial-and-error experience to produce robot-feasible motions.
To mitigate this limitation, we consider two directions for improving the feedback: (i) directly constraining infeasible motions during demonstration and (ii) explicitly indicating the degree of improvement across demonstrations.

For direction (i), the hand-mounted demonstration interface could incorporate an arm-worn exoskeleton frame that applies resistance or constraints to motions that are difficult for the robot to execute, helping the subject to intuitively understand the feasibility for each of the six-dimensional end-effector pose components~\cite{wu2024gello, fang2023low}.
For direction (ii), we could compare the feasibility of the most recent demonstrated motion with the feasibilities of previously collected demonstrated motions to indicate whether the motion has improved, helping subjects to refine their motions more efficiently.

\subsection{Constraints and improvements for dynamics learning}
A key practical challenge in FABCO is the cost of collecting robot motion data for learning the forward and inverse dynamics models.
Because feasibility evaluation and action estimation rely on these models, achieving sufficient accuracy can require operating a real robot for long periods and across a wide range of motions to gather sufficient data.
Reducing this data collection time while maintaining dynamics accuracy is thus an important direction for improvement.

One approach for learning a dynamics model more efficiently is to incorporate robot kinematics as prior knowledge into the model.
Concretely, forward kinematics and the Jacobian could be used to constrain the model inputs and outputs, embedding structures such as rigid-body motion and joint limits into the dynamics model so that a consistent approximation can be obtained even with limited data~\cite{lutter2019deep,wong2022oscar}.
Another possible approach is to carry out offline pretraining in simulation and then fine-tune the model using a small amount of real-world data~\cite{li2023offline}.
In addition, supplementing the method with an active data-collection strategy---for example, prioritizing exploration of regions with sparse sampling density based on model uncertainty---could improve data efficiency compared with uniform and uninformed exploration of the workspace.
By combining these directions, it might be possible to both reduce the real-robot time required for dynamics learning and improve model accuracy, thus increasing the reliability of feedback and potentially improving policy performance as well.

\subsection{Limitations of FABCO and possible improvements}
In this study, we built an imitation learning framework based on behavior cloning (BC), where a policy is learned after collecting demonstration data.
While BC enables a simple and practical learning pipeline, it is prone to performance degradation due to compounding errors, and FABCO shares this limitation.
A possible way to mitigate this issue is to incorporate interactive imitation learning (IIL) into FABCO.

IIL allows a demonstrator to intervene during robot execution and sequentially incorporates the resulting information into the training data to improve the policy.
Representative approaches include DART, which improves robustness by collecting demonstrations under injected perturbations~\cite{laskey2017, tahara2022, tahara2023, oh2023, Oh_2024}, and DAgger, in which the demonstrator provides corrective actions for states visited by the learner to prevent task failures~\cite{Ross2010ARO, michael2016, michael2019, kunal2019}.
In our setting, similar to DAgger, we could use the FDM and IDM to simulate policy rollouts and identify end-effector poses that are likely to lead to task failure.
By asking the demonstrator to initiate a demonstration from such failure-prone poses, we could collect demonstration data more efficiently and more directly improve policy performance.
Therefore, integrating an IIL framework into FABCO may create a closed-loop data collection and learning cycle that improves imitation performance more directly and continuously.

In addition, future work should consider the feasibility of gripper opening and closing.
Due to differences in dynamics between the hand-mounted demonstration interface and the robot gripper, demonstrated opening/closing motions may not be feasible for the robot to execute as intended.
In our experiments, it was difficult to provide feasibility feedback for both end-effector motion and finger opening and closing while keeping the feedback easy to interpret.
Therefore, we focused on the end-effector motion's feasibility.
In future work, we will improve the interface by introducing a mechanism that provides feasibility-dependent force resistance to the fingers during demonstration.
Furthermore, by leveraging a finger dynamics model to estimate the feasibility of demonstrated gripper motions and applying force resistance based on the estimated feasibility, we aim to build a system that permits imitation of entire demonstrated motions, including gripper opening and closing.

%%%%%%%%%%%%%%%%%%%%%%%%%%%%%%%%%%%%%%%%%%%%%%%%%%%%%%%%%%%%%%%%%%%%%%%%%%%%%%%%
%%%%%%%%%%%%%%%%%%%%%%%%%%%%%%%%%%%%%%%%%%%%%%%%%%%%%%%%%%%%%%%%%%%%%%%%%%%%%%%%
%%%%%%%%%%%%%%%%%%%%%%%%%%%%%%%%%%%%%%%%%%%%%%%%%%%%%%%%%%%%%%%%%%%%%%%%%%%%%%%%
% \section{Conclusion}
% In this study, we proposed FABCO, an imitation learning framework that provides feasibility feedback during motion provision to encourage demonstrators to provide robot-feasible demonstrated motions.
% To compare feasibility feedback modalities and to evaluate the effectiveness of feasibility-aware policy learning, we conducted human-subject studies with 15 subjects on two tasks: the peg insertion task and the circle tracing task.
% The results confirmed that introducing feasibility feedback improves policy performance by more than 3.2 times compared with the No FB demonstration, that the most effective feedback modality depends on task characteristics, and that feasibility-aware policy learning improves policy performance particularly when the amount of demonstration data are limited.
% In addition, the questionnaire results suggest that visual feedback fosters a stronger sense of task accomplishment among demonstrator, haptic feedback is often preferred because it is easy to understand and supports on-the-spot refinement, and visuo-haptic feedback has the potential to combine the advantages of both modalities.

\section{Conclusion}
In this study, we proposed FABCO, an imitation learning framework that provides feasibility feedback during motion performance to encourage demonstrators to provide robot-feasible motions.
To compare feasibility feedback modalities and to evaluate the effectiveness of feasibility-aware policy learning, we conducted a human-subject experiment with 15 subjects on two tasks: a peg-insertion task and a circle-tracing task.
The results confirm that introducing feasibility feedback improves policy performance by more than 3.2 times compared with a No FB demonstration, that the most effective feedback modality depends on task characteristics, and that feasibility-aware policy learning improves policy performance, particularly when the amount of demonstration data is limited.
In addition, the questionnaire results suggest that visual feedback fosters a stronger sense of task accomplishment among demonstrators, that haptic feedback is often preferred because it is easy to understand and supports on-the-spot motion refinement, and that visuo-haptic feedback offers much promise by combining the advantages of both modalities.

%%%%%%%%%%%%%%%%%%%%%%%%%%%%%%%%%%%%%%%%%%%%%%%%%%%%%%%%%%%%%%%%%%%%%%%%%%%%%%%%

% \vfill
\bibliographystyle{IEEEtran}
\bibliography{reference}

\begin{thebibliography}{10}
\providecommand{\url}[1]{#1}
\csname url@samestyle\endcsname
\providecommand{\newblock}{\relax}
\providecommand{\bibinfo}[2]{#2}
\providecommand{\BIBentrySTDinterwordspacing}{\spaceskip=0pt\relax}
\providecommand{\BIBentryALTinterwordstretchfactor}{4}
\providecommand{\BIBentryALTinterwordspacing}{\spaceskip=\fontdimen2\font plus
\BIBentryALTinterwordstretchfactor\fontdimen3\font minus
  \fontdimen4\font\relax}
\providecommand{\BIBforeignlanguage}[2]{{%
\expandafter\ifx\csname l@#1\endcsname\relax
\typeout{** WARNING: IEEEtran.bst: No hyphenation pattern has been}%
\typeout{** loaded for the language `#1'. Using the pattern for}%
\typeout{** the default language instead.}%
\else
\language=\csname l@#1\endcsname
\fi
#2}}
\providecommand{\BIBdecl}{\relax}
\BIBdecl

\bibitem{osa2018}
T.~Osa, J.~Pajarinen, G.~Neumann, J.~A. Bagnell, P.~Abbeel, and J.~Peters, ``An
  algorithmic perspective on imitation learning,'' \emph{Foundations and Trends
  in Robotics}, vol.~7, no. 1-2, pp. 1--179, 2018.

\bibitem{chi2024diffusionpolicy}
C.~Chi, Z.~Xu, S.~Feng, E.~Cousineau, Y.~Du, B.~Burchfiel, R.~Tedrake, and
  S.~Song, ``Diffusion policy: Visuomotor policy learning via action
  diffusion,'' \emph{The International Journal of Robotics Research}, vol.~44,
  pp. 1684--1704, 2024.

\bibitem{janner2022diffuser}
M.~Janner, Y.~Du, J.~Tenenbaum, and S.~Levine, ``Planning with diffusion for
  flexible behavior synthesis,'' in \emph{International Conference on Machine
  Learning (ICML)}, 2022, pp. 9902--9915.

\bibitem{bharadhwaj2024roboagent}
H.~Bharadhwaj, J.~Vakil, M.~Sharma, A.~Gupta, S.~Tulsiani, and V.~Kumar,
  ``Roboagent: Generalization and efficiency in robot manipulation via semantic
  augmentations and action chunking,'' in \emph{International Conference on
  Robotics and Automation (ICRA)}, 2024, pp. 4788--4795.

\bibitem{schaal1996learning}
S.~Schaal, ``Learning from demonstration,'' in \emph{Conference on neural
  information processing systems(NeurIPS)}, 1996, pp. 1040--1046.

\bibitem{argall2009survey}
B.~D. Argall, S.~Chernova, M.~Veloso, and B.~Browning, ``A survey of robot
  learning from demonstration,'' \emph{Robotics and autonomous systems},
  vol.~57, no.~5, pp. 469--483, 2009.

\bibitem{torabi2018}
F.~Torabi, G.~Warnell, and P.~Stone, ``Behavioral cloning from observation,''
  in \emph{International Joint Conference on Artificial Intelligence (IJCAI)},
  2018, pp. 4950--4957.

\bibitem{baker2022video}
B.~Baker, I.~Akkaya, P.~Zhokov, J.~Huizinga, J.~Tang, A.~Ecoffet, B.~Houghton,
  R.~Sampedro, and J.~Clune, ``Video pretraining (vpt): Learning to act by
  watching unlabeled online videos,'' in \emph{Conference on Neural Information
  Processing Systems (NeurIPS)}, 2022, pp. 24\,639--24\,654.

\bibitem{torabi2019adversarial}
F.~Torabi, G.~Warnell, and P.~Stone, ``Adversarial imitation learning from
  state-only demonstrations,'' in \emph{Proceedings of the 18th International
  Conference on Autonomous Agents and MultiAgent Systems}, 2019, pp.
  2229--2231.

\bibitem{WangSWZFL24}
C.~Wang, H.~Shi, W.~Wang, R.~Zhang, L.~Fei-Fei, and C.~K. Liu, ``Dexcap:
  Scalable and portable mocap data collection system for dexterous
  manipulation,'' in \emph{Robotics: Science and Systems(RSS)}, 2024.

\bibitem{chen2024arcap}
S.~Chen, C.~Wang, K.~Nguyen, L.~Fei-Fei, and C.~K. Liu, ``Arcap: Collecting
  high-quality human demonstrations for robot learning with augmented reality
  feedback,'' in \emph{International Conference on Robotics and Automation
  (ICRA)}, 2025, pp. 8291--8298.

\bibitem{chi2024universal}
C.~Chi, Z.~Xu, C.~Pan, E.~Cousineau, B.~Burchfiel, S.~Feng, R.~Tedrake, and
  S.~Song, ``Universal manipulation interface: In-the-wild robot teaching
  without in-the-wild robots,'' in \emph{Robotics: Science and Systems (RSS)},
  2024.

\bibitem{hamaya2020}
M.~Hamaya, F.~von Drigalski, T.~Matsubara, K.~Tanaka, R.~Lee, C.~Nakashima,
  Y.~Shibata, and Y.~Ijiri, ``Learning soft robotic assembly strategies from
  successful and failed demonstrations,'' in \emph{International Conference on
  Intelligent Robots and Systems (IROS)}, 2020, pp. 8309--8315.

\bibitem{young2021}
S.~Young, D.~Gandhi, S.~Tulsiani, A.~Gupta, P.~Abbeel, and L.~Pinto, ``Visual
  {{Imitation Made Easy}},'' in \emph{{{Conference}} on {{Robot Learning}}
  (CoRL)}, 2021, pp. 1992--2005.

\bibitem{legato}
M.~Seo, H.~A. Park, S.~Yuan, Y.~Zhu, and L.~Sentis, ``Legato: Cross-embodiment
  imitation using a grasping tool,'' \emph{IEEE Robotics and Automation
  Letters}, vol.~10, no.~3, pp. 2854--2861, 2025.

\bibitem{pmlr-v270-ha25a}
H.~Ha, Y.~Gao, Z.~Fu, J.~Tan, and S.~Song, ``Umi-on-legs: Making manipulation
  policies mobile with manipulation-centric whole-body controllers,'' in
  \emph{Conference on Robot Learning (CoRL)}, 2025, pp. 5254--5270.

\bibitem{zeng2025activeumiroboticmanipulationactive}
Q.~Zeng, C.~Li, J.~S. John, Z.~Zhou, J.~Wen, G.~Feng, Y.~Zhu, and Y.~Xu,
  ``Activeumi: Robotic manipulation with active perception from robot-free
  human demonstrations,'' 2025.

\bibitem{wong2022error}
J.~Wong, A.~Tung, A.~Kurenkov, A.~Mandlekar, L.~Fei-Fei, S.~Savarese, and
  R.~Mart{\'\i}n-Mart{\'\i}n, ``Error-aware imitation learning from
  teleoperation data for mobile manipulation,'' in \emph{Conference on Robot
  Learning (CoRL)}, 2022, pp. 1367--1378.

\bibitem{fu2024mobile}
Z.~Fu, T.~Z. Zhao, and C.~Finn, ``Mobile aloha: Learning bimanual mobile
  manipulation with low-cost whole-body teleoperation,'' in \emph{{Conference
  on Robot Learning (CoRL)}}, 2025, pp. 4066--4083.

\bibitem{pmlr-v270-zhao25b}
T.~Z. Zhao, J.~Tompson, D.~Driess, P.~Florence, S.~K.~S. Ghasemipour, C.~Finn,
  and A.~Wahid, ``Aloha unleashed: A simple recipe for robot dexterity,'' in
  \emph{Conference on Robot Learning (CoRL)}, 2025, pp. 1910--1924.

\bibitem{cheng2024opentelevision}
X.~Cheng, J.~Li, S.~Yang, G.~Yang, and X.~Wang, ``Open-television:
  Teleoperation with immersive active visual feedback,'' in \emph{Conference on
  Robot Learning (CoRL)}, 2024, pp. 2729--2749.

\bibitem{cuan2024leveraginghapticfeedbackimprove}
C.~Cuan, A.~Okamura, and M.~Khansari, ``Leveraging haptic feedback to improve
  data quality and quantity for deep imitation learning models,'' \emph{IEEE
  Transactions on Haptics}, vol.~17, no.~4, pp. 984--991, 2024.

\bibitem{Zhao-RSS-23}
T.~Z. Zhao, V.~Kumar, S.~Levine, and C.~Finn, ``{Learning Fine-Grained Bimanual
  Manipulation with Low-Cost Hardware},'' in \emph{Robotics: Science and
  Systems (RSS)}, 2023.

\bibitem{takahashi2025feasibilityawareimitationlearningobservations}
K.~Takahashi, H.~Sasaki, and T.~Matsubara, ``Feasibility-aware imitation
  learning from observations through a hand-mounted demonstration interface,''
  in \emph{International Conference on Robotics and Automation (ICRA)}, 2025,
  pp. 7822--7828.

\bibitem{HART1988139}
S.~G. Hart and L.~E. Staveland, ``Development of nasa-tlx (task load index):
  Results of empirical and theoretical research,'' in \emph{Human Mental
  Workload}, 1988, vol.~52, pp. 139--183.

\bibitem{liu2018imitation}
Y.~Liu, A.~Gupta, P.~Abbeel, and S.~Levine, ``Imitation from observation:
  Learning to imitate behaviors from raw video via context translation,'' in
  \emph{International Conference on Robotics and Automation (ICRA)}, 2018, pp.
  1118--1125.

\bibitem{smith2019avid}
L.~Smith, N.~Dhawan, M.~Zhang, P.~Abbeel, and S.~Levine, ``Avid: Learning
  multi-stage tasks via pixel-level translation of human videos,'' in
  \emph{Robotics: Science and Systems (RSS)}, 2020.

\bibitem{qin2021from}
Y.~Qin, H.~Su, and X.~Wang, ``From one hand to multiple hands: Imitation
  learning for dexterous manipulation from single-camera teleoperation,''
  \emph{IEEE Robotics and Automation Letters}, vol.~7, no.~4, pp.
  10\,873--10\,881, 2022.

\bibitem{zhang2022learning}
Q.~Zhang, Z.~Peng, and B.~Zhou, ``Learning to drive by watching youtube videos:
  Action-conditioned contrastive policy pretraining,'' in \emph{European
  Conference on Computer Vision (ECCV)}, 2022, pp. 111--128.

\bibitem{bahl2022human}
S.~Bahl, A.~Gupta, and D.~Pathak, ``Human-to-robot imitation in the wild,'' in
  \emph{Robotics: Science and Systems (RSS)}, 2022.

\bibitem{karnan2022voila}
H.~Karnan, G.~Warnell, X.~Xiao, and P.~Stone, ``Voila: Visual-observation-only
  imitation learning for autonomous navigation,'' in \emph{International
  Conference on Robotics and Automation (ICRA)}, 2022, pp. 2497--2503.

\bibitem{sikchi2024dual}
H.~Sikchi, C.~Chuck, A.~Zhang, and S.~Niekum, ``A dual approach to imitation
  learning from observations with offline datasets,'' in \emph{Conference on
  Robot Learning (CoRL)}, 2025, pp. 1125--1147.

\bibitem{sukkar2023guided}
F.~Sukkar, V.~H. Moreno, T.~Vidal-Calleja, and J.~Deuse, ``Guided learning from
  demonstration for robust transferability,'' in \emph{International Conference
  on Robotics and Automation (ICRA)}, 2023, pp. 5048--5054.

\bibitem{cao2021corl}
Z.~Cao, Y.~Hao, M.~Li, and D.~Sadigh, ``Learning feasibility to imitate
  demonstrators with different dynamics,'' in \emph{Conference on Robot
  Learning (CoRL)}, 2022, pp. 363--372.

\bibitem{cao2021ral}
Z.~Cao and D.~Sadigh, ``Learning from imperfect demonstrations from agents with
  varying dynamics,'' \emph{IEEE Robotics and Automation Letters}, vol.~6,
  no.~3, pp. 5231--5238, 2021.

\bibitem{Betz_2021}
T.~Betz, H.~Fujiishi, and T.~Kobayashi, ``Behavioral cloning from observation
  with bi-directional dynamics model,'' in \emph{International Symposium on
  System Integration (SII)}, 2021, pp. 184--189.

\bibitem{jungeblut2022complexityhausdorffdistance}
P.~Jungeblut, L.~Kleist, and T.~Miltzow, ``The complexity of the hausdorff
  distance,'' \emph{Discrete \& Computational Geometry}, vol.~71, no.~1, pp.
  177--213, 2024.

\bibitem{wu2024gello}
P.~Wu, Y.~Shentu, Z.~Yi, X.~Lin, and P.~Abbeel, ``Gello: A general, low-cost,
  and intuitive teleoperation framework for robot manipulators,'' in
  \emph{International Conference on Intelligent Robots and Systems (IROS)},
  2024, pp. 12\,156--12\,163.

\bibitem{fang2023low}
H.~Y. Fang, {Hongjie and Fang}, J.~Ren, J.~Chen, R.~Zhang, W.~Wang, and C.~Lu,
  ``Low-cost exoskeletons for learning whole-arm manipulation in the wild,'' in
  \emph{Conference on Robot Learning (CoRL)}, 2023, pp. 1121--1133.

\bibitem{lutter2019deep}
M.~Lutter, K.~Listmann, and J.~Peters, ``Deep lagrangian networks for
  end-to-end learning of energy-based control for under-actuated systems,'' in
  \emph{International Conference on Intelligent Robots and Systems (IROS)},
  2019, pp. 7718--7725.

\bibitem{wong2022oscar}
J.~Wong, V.~Makoviychuk, A.~Anandkumar, and Y.~Zhu, ``Oscar: Data-driven
  operational space control for adaptive and robust robot manipulation,'' in
  \emph{International Conference on Robotics and Automation (ICRA)}, 2022, pp.
  10\,519--10\,526.

\bibitem{li2023offline}
X.~Li, W.~Shang, and S.~Cong, ``Offline reinforcement learning of robotic
  control using deep kinematics and dynamics,'' \emph{IEEE/ASME Transactions on
  Mechatronics}, vol.~29, no.~4, pp. 2428--2439, 2023.

\bibitem{laskey2017}
M.~Laskey, J.~Lee, R.~Fox, A.~Dragan, and K.~Goldberg, ``Dart: Noise injection
  for robust imitation learning,'' in \emph{Conference on Robot Learning
  (CoRL)}, 2017, pp. 143--156.

\bibitem{tahara2022}
H.~Tahara, H.~Sasaki, H.~Oh, B.~Michael, and T.~Matsubara,
  ``Disturbance-injected robust imitation learning with task achievement,'' in
  \emph{International Conference on Robotics and Automation (ICRA)}, 2022, pp.
  2466--2472.

\bibitem{tahara2023}
H.~Tahara, H.~Sasaki, H.~Oh, E.~Anarossi, and T.~Matsubara, ``Disturbance
  injection under partial automation: Robust imitation learning for
  long-horizon tasks,'' \emph{IEEE Robotics and Automation Letters}, vol.~8,
  no.~5, pp. 2724--2731, 2023.

\bibitem{oh2023}
H.~Oh, H.~Sasaki, B.~Michael, and T.~Matsubara, ``Bayesian disturbance
  injection: Robust imitation learning of flexible policies for robot
  manipulation,'' \emph{Neural Networks}, vol. 158, pp. 42--58, 2023.

\bibitem{Oh_2024}
H.~Oh and T.~Matsubara, ``Leveraging demonstrator-perceived precision for safe
  interactive imitation learning of clearance-limited tasks,'' \emph{IEEE
  Robotics and Automation Letters}, vol.~9, no.~4, p. 3387^^e2^^80^^933394,
  2024.

\bibitem{Ross2010ARO}
S.~Ross, G.~J. Gordon, and J.~A. Bagnell, ``A reduction of imitation learning
  and structured prediction to no-regret online learning,'' in
  \emph{International Conference on Artificial Intelligence and Statistics
  (AISTATS)}, 2011, pp. 627--635.

\bibitem{michael2016}
M.~Laskey, S.~Staszak, W.~Y. Hsieh, J.~Mahler, F.~T. Pokorny, A.~D. Dragan, and
  K.~Goldberg, ``Shiv: Reducing supervisor burden in dagger using support
  vectors for efficient learning from demonstrations in high dimensional state
  spaces,'' in \emph{International Conference on Robotics and Automation
  (ICRA)}, 2016, pp. 462--469.

\bibitem{michael2019}
M.~Kelly, C.~Sidrane, K.~Driggs-Campbell, and M.~J. Kochenderfer, ``Hg-dagger:
  Interactive imitation learning with human experts,'' in \emph{International
  Conference on Robotics and Automation (ICRA)}, 2019, pp. 8077--8083.

\bibitem{kunal2019}
K.~Menda, K.~Driggs-Campbell, and M.~J. Kochenderfer, ``Ensembledagger: A
  bayesian approach to safe imitation learning,'' in \emph{International
  Conference on Intelligent Robots and Systems (IROS)}, 2019, pp. 5041--5048.

\end{thebibliography}

\end{document}